\documentclass{article} 
\usepackage{iclr2024_conference,times}

\usepackage{multirow} 
\usepackage{graphicx} 

\usepackage[utf8]{inputenc}
\usepackage{amsmath}
\usepackage{amssymb}
\usepackage{amsthm}
\usepackage{algpseudocode} 
\usepackage{algorithm} 
\usepackage{subcaption} 
\usepackage[export]{adjustbox} 

\usepackage{enumitem}

\usepackage{amsmath,amsfonts,bm}

\def\eqref#1{equation~\ref{#1}}

\def\1{\bm{1}}

\def\mI{{\bm{I}}}

\def\mK{{\bm{K}}}

\def\mU{{\bm{U}}}

\def\mX{{\bm{X}}}
\def\mY{{\bm{Y}}}

\DeclareMathAlphabet{\mathsfit}{\encodingdefault}{\sfdefault}{m}{sl}
\SetMathAlphabet{\mathsfit}{bold}{\encodingdefault}{\sfdefault}{bx}{n}

\newcommand{\E}{\mathbb{E}}

\newcommand{\R}{\mathbb{R}}

\DeclareMathOperator*{\argmin}{arg\,min}

\usepackage{hyperref}
\usepackage{url}
\usepackage{comment}

\newtheorem{innercustomgeneric}{\customgenericname}
\providecommand{\customgenericname}{}
\newcommand{\newcustomtheorem}[2]{%
  \newenvironment{#1}[1]
  {%
   \renewcommand\customgenericname{#2}%
   \renewcommand\theinnercustomgeneric{##1}%
   \innercustomgeneric
  }
  {\endinnercustomgeneric}
}

\newcustomtheorem{customthm}{Theorem}
\newcustomtheorem{customlemma}{Lemma}

\newtheorem{definition}{Definition}
\newtheorem{theorem}{Theorem}
\newtheorem{lemma}{Lemma}

\usepackage{color}

\graphicspath{{./fig/}} 

\title{Rethinking Backdoor Attacks on Dataset Distillation: A Kernel Method Perspective}

\author{Ming-Yu Chung \\
National Taiwan University\\
\And
Sheng-Yen Chou \\
The Chinese University of Hong Kong \\
\And
Chia-Mu Yu \\
National Yang Ming Chiao Tung University \\
\And
Pin-Yu Chen \\
IBM Research \\
\And
Sy-Yen Kuo \\
National Taiwan University\\
\And
Tsung-Yi Ho \\
The Chinese University of Hong Kong \\
}

\iclrfinalcopy 
\begin{document}

\maketitle

\begin{abstract}
Dataset distillation offers a potential means to enhance data efficiency in deep learning. Recent studies have shown its ability to counteract backdoor risks present in original training samples. In this study, we delve into the theoretical aspects of backdoor attacks and dataset distillation based on kernel methods. We introduce two new theory-driven trigger pattern generation methods specialized for dataset distillation.
Following a comprehensive set of analyses and experiments, we show that our optimization-based trigger design framework informs effective backdoor attacks on dataset distillation. Notably, datasets poisoned by our designed trigger prove resilient against conventional backdoor attack detection and mitigation methods.  Our empirical results validate that the triggers developed using our approaches are proficient at executing resilient backdoor attacks. 
\footnote{Code is available at \url{https://github.com/Mick048/KIP-based-backdoor-attack.git}.}
\end{abstract}

\section{Introduction}\label{sec: Introduction}
In recent years, deep neural networks have achieved significant success in many fields, such as natural language modeling, computer vision, medical diagnosis, etc. These successes are usually built on large-scale datasets consisting of millions or even billions of samples. Under this scale of datasets, training a model becomes troublesome because of the need for sufficiently large memory to store the datasets or the need for special infrastructure to train a model. To deal with this problem, dataset distillation~\citep{wang2018dataset} or dataset condensation~\citep{ zhao2020dataset} is designed to compress the information of large datasets into a small synthetic dataset. These small datasets generated by dataset distillation, called distilled datasets, still retain a certain degree of utility. Under the same model (neural network) structure, the performance of the model trained on the distilled dataset is only slightly lower than that of the model trained on the original large-scale dataset.

However, with the development of dataset distillation techniques, the related security and privacy issues started to emerge \citep{liu2023copyright, liu2023backdoor, dong2022privacy}. In this paper, we focus on backdoor attacks on dataset distillation. In particular, as each distilled sample does not have a clear connection to the original samples, a straightforward stealthy backdoor attack is to poison a benign dataset first and then derive the corresponding distilled poisoned dataset. One can expect that the triggers can hardly be detected visually in the distilled poisoned dataset. However, these triggers, if not designed properly, can be diluted during dataset distillation, making backdoor attacks ineffective. 

\citet{liu2023backdoor} empirically demonstrate the feasibility of generating a poisoned dataset surviving dataset distillation. In particular, \citet{liu2023backdoor} propose DOORPING as a distillation-resilient backdoor. However, DOORPING suffers from two major weaknesses. First, the resiliency and optimality of a backdoor against dataset distillation remain unclear, mainly due to the lack of a theoretical foundation for the distillation resiliency. Second, DOORPING relies on a bi-level optimization and, as a consequence, consumes a significant amount of time to generate backdoor triggers.

To bridge this gap, this paper makes a step toward dataset distillation-resilient backdoors with a theoretical foundation. Our contributions can be summarized as follows:
\begin{itemize}[leftmargin=*]
    \item To the best of our knowledge, we establish the first theoretical framework to characterize backdoor effects on dataset distillation, which explains why certain backdoors survive dataset distillation.
    \item 
    We propose two theory-induced backdoors, \textsf{simple-trigger} and \textsf{relax-trigger}. In particular, \textsf{relax-trigger} and DOORPING share the same clean test accuracy (CTA) and attack success rate (ASR). However, \textsf{relax-trigger} relies only on ordinary (single-level) optimization procedures and can be computationally efficient.

    \item We experimentally show both \textsf{simple-trigger} and \textsf{relax-trigger} signify the advanced threat vector to either completely break or weaken eight existing defenses. In particular, \textsf{relax-trigger} can evade all eight existing backdoor detection and cleansing methods considered in this paper.
\end{itemize}

\section{Background and Related Works}\label{sec:2}
\paragraph{Dataset Distillation.}
Dataset distillation is a technique for compressing the information of a target dataset into a small synthetic dataset. The explicit definition can be described as follows. Consider the input space $\mathcal{X}\subset \mathbb{R}^{d}$, the label space $\mathcal{Y}\subset \mathbb{R}^{C}$, and the distribution $(x, y)\sim \mathcal{D}$, where $x\in\mathcal{X}$ and $y\in\mathcal{Y}$. Suppose we are given a dataset denoted by $\mathcal{T} = \{(x_t, y_t)\}^{N}_{t=1} \sim \mathcal{D}^N$ where $x_t \in \mathcal{X}$, $y_t \in \mathcal{Y}$, and $N$ is the number of samples, and a synthetic dataset denoted as $\mathcal{S} = \{(x_s, y_s)\}_{s=1}^{N_{\mathcal{S}}}$ where $x_s \in \mathcal{X}$, $y_s \in \mathcal{Y}$, $N_{\mathcal{S}}$ is the number of samples in $\mathcal{S}$, and $N_{\mathcal{S}}\ll N$. The synthetic dataset $\mathcal{S}^*$ generated by a dataset distillation method can be formulated as 
\begin{equation}\label{intro:dd}
    \mathcal{S}^* = \argmin_{\mathcal{S}} \mathcal{L}(\mathcal{S}, \mathcal{T}),
\end{equation}
where $\mathcal{L}$ is some function to measure the information loss between $\mathcal{S}$ and $\mathcal{T}$. There are several types of $\mathcal{L}$. One of the most straightforward ways to define $\mathcal{L}$ is to measure the model's performance. In this sense, the dataset distillation can be reformulated as 
\begin{align}\label{def:dd}
    \mathcal{S}^* 
    = 
    \argmin_{\mathcal{S}} \frac{1}{N}\ell(f_{\mathcal{S}}, \mathcal{T})
    \text{ subject to } 
    f_{\mathcal{S}} = \argmin_{f\in\mathcal{H}} \frac{1}{N_{\mathcal{S}}}\ell(f, \mathcal{S}) + \lambda\|f\|^2_{\mathcal{H}}
\end{align}
where the model (a classifier) is denoted as $f:\mathcal{X}\rightarrow\mathcal{Y}$, $\mathcal{H}$ is some collection of models (hypothesis class), $\ell$ is the loss function measuring the loss of model evaluated on the dataset, $\lambda \geq0$ is the weight for the regularization term, and $\|\|_{\mathcal{H}}$ is some norm defined on $\mathcal{H}$. Eq.~(\ref{def:dd})  forms a bi-level optimization problem. This type of dataset distillation is categorized as \textit{performance-matching dataset distillation} in \citep{yu2023dataset}. For example, all of the methods from \citep{wang2018dataset, nguyen2020dataset, loo2022efficient, zhou2022dataset, loo2023dataset} are performance-matching dataset distillation, while the methods from \citep{zhao2023dataset, lee2022dataset, wang2022cafe, zhao2020dataset, lee2022dataset2, liu2022dataset, liu2023dream, wang2023dim} belong to either parameter-preserving or distribution-preserving. In this paper, we focus only on performance-matching dataset distillation, with a particular example on kernel inducing points (KIP) from \cite{nguyen2020dataset}. 

\paragraph{Reproducing Kernel Hilbert Space and KIP.}
In general, the inner optimization problem in Eq.~(\ref{def:dd}) does not have a closed-form solution, which not only increases the computational cost, but also increases the difficulty of analyzing this problem. To alleviate this problem, we assume our model lies in the reproducing kernel Hilbert space (RKHS) \citep{aronszajn1950theory, berlinet2011reproducing, ghojogh2021reproducing}. 

\begin{definition}[Kernel]\label{def:kernel}
$k:\mathcal{X}\times\mathcal{X}\rightarrow\R$ is a kerenl if the following two points hold.
    (a) $\forall x, x' \in\mathcal{X}$, the kernel $k$ is symmetric; i.e., $k(x, x') = k(x', x)$.
    (b) $\forall n\in\mathbb{N}$, $\forall \{x_1, x_2, \dots, x_n\}$ where each $x_i$ are sampled from $\mathcal{X}$, the kernel matrix $\mK$ defined as $\mK_{ij} := k(x_i, x_j)$ is postive semi-definite.
\end{definition}

\begin{definition}[Reproducing Kernel Hilbert Space]\label{def:rkhs}
Given an kernel $k:\mathcal{X}\times\mathcal{X}\rightarrow\mathbb{R}$, the collection of real-valued model $\mathcal{H}_{k}=\{f:\mathcal{X}\rightarrow\mathbb{R}\}$ is a reproducing kernel Hilbert space corresponding to the kernel $k$, if
    (a) $\mathcal{H}_k$ is a Hilbert space corresponding to the inner product $\langle\cdot, \cdot\rangle_{\mathcal{H}_k}$,
    (b) $\forall x\in\mathcal{X}$, ~$k(\cdot, x) \in \mathcal{H}_{k}$,
    (c) $\forall x\in\mathcal{X}$ and $f\in\mathcal{H}_{k}$, $f(x) = \langle f, k(\cdot, x)\rangle_{\mathcal{H}_k}$ (Reproducing property).
\end{definition}

There are several advantages to considering RKHS for solving optimization problems. One of the most beneficial properties is that there is Representer Theorem \citep{kimeldorf1971some,ghojogh2021reproducing} induced by the reproducing property. In particular, consider the optimization problem:
\begin{equation}
    f^* = \argmin_{f\in\mathcal{H}_k} \frac{1}{N}\sum_{i=1}^N\ell(f(x_i), y_i) + \lambda\|f\|^2_{\mathcal{H}_k},
\end{equation}
where $f:\mathcal{X}\rightarrow\R$, $y_i \in \R$, $\lambda\geq 0$ is the weight for the regularization term.
The solution of the optimization problem $f^*$ can be expressed as the linear combination of $\{k(\cdot, x_i)\}_i^N$. Furthermore, if we set $\ell(f, (x, y)) = \|f(x)-y\|^2_2$, there is a closed-form expression for $f^*$:
\begin{equation}\label{sol:closed-form}
    f^*(x) = k(x, \mX)[k(\mX, \mX) + N\lambda \mI]^{-1} \mY,
\end{equation}
where $k(x, \mX) = [k(x, x_1), k(x,x_2), \dots, k(x, x_N)]$, $k(\mX, \mX)$ is an $N\times N$ matrix with $[k(\mX, \mX)]_{ij} = k(x_i, x_j)$, and $\mY = [y_1, y_2, \dots, y_N]^T$. Now, we return to Eq.~(\ref{def:dd}). By rewriting the model $f: \mathcal{X} \rightarrow \mathcal{Y}\subset\R^c$ as $[f^1, f^2, \cdots, f^c]^T$, where each $f^i: \mathcal{X}\rightarrow\R$ is a real-valued function and $f^i$ is bounded in the RKHS $\mathcal{H}_k$, the inner optimization problem for $f_{\mathcal{S}}$ in Eq.~(\ref{def:dd}) can be considered as $c$ independent optimization problems and each problem has a closed-form solution as shown in Eq.~(\ref{sol:closed-form}).
Thus, the solution of the inner optimization problem can be expressed as
\begin{equation}\label{sol:c-closed-form}
    f_{\mathcal{S}}(x)^T = k(x, \mX_{\mathcal{S}}) [k(\mX_{\mathcal{S}}, \mX_{\mathcal{S}}) + N_{\mathcal{S}}\lambda\mI]^{-1} \mY_{\mathcal{S}}, 
\end{equation}
where $k(x, \mX_{\mathcal{S}}) = [k(x, x_{s_1}), k(x,x_{s_2}), \dots, k(x, x_{s_{N_{\mathcal{S}}}})]$, $k(\mX_{\mathcal{S}}, \mX_{\mathcal{S}})$ is a $N_{\mathcal{S}}\times N_{\mathcal{S}}$ matrix with $[k(\mX_{\mathcal{S}}, \mX_{\mathcal{S}})]_{ij} = k(x_{s_i}, x_{s_j})$, and $\mY_{\mathcal{S}}$ is a $N_{\mathcal{S}}\times c$ matrix with $\mY_{\mathcal{S}} = [y_{s_1}, y_{s_2}, \dots, y_{s_{N_{\mathcal{S}}}}]^T$.

Then, the dataset distillation problem can be expressed as
\begin{equation}\label{KIP}
    \mathcal{S}^* = \argmin_{\mathcal{S}} \frac{1}{N}\sum_{t=1}^{N} \|f_\mathcal{S}(x_t) - y_t\|^2_2,
\end{equation}
where $f_\mathcal{S}(x)^T = k(x, \mX_{\mathcal{S}}) [k(\mX_{\mathcal{S}}, \mX_{\mathcal{S}}) + N_{\mathcal{S}}\lambda\mI]^{-1} \mY_{\mathcal{S}}$ as shown in Eq.~(\ref{sol:c-closed-form}). We reduce a two-level optimization problem to a one-level optimization problem using RKHS. Essentially, KIP~\citep{nguyen2020dataset} can be formulated as Eq.~(\ref{KIP}).

An important problem for Eq.~(\ref{KIP}) is how to construct or select a kernel $k(\cdot, \cdot)$. Nevertheless, we do not discuss this problem in this paper. We directly consider the neural tangent kernel (NTK) \citep{jacot2018neural, he2020bayesian, lee2019wide} induced by a three-layer neural network as the kernel $k(\cdot, \cdot)$ to do the experiment in Section~\ref{sec: Experiment}. 

\paragraph{Backdoor Attack.}\label{sec:2.3}
Backdoor attack introduces some malicious behavior into the model without degrading the model's performance on the original task by poisoning the dataset~\citep{gu2017badnets, chen2017targeted, liu2018trojaning, turner2019label, nguyen2020input, barni2019new,li2021invisible,nguyen2021wanet,liu2020reflection,tang2021demon,qi2022revisiting,souri2022sleeper}. To be more specific, consider the following scenario. Suppose there are two types of distributions, $(x_a, y_a)\sim\mathcal{D}_A$ and $(x_b, y_b)\sim\mathcal{D}_B$. $\mathcal{D}_A$ corresponds to the original normal behavior, while $\mathcal{D}_B$ corresponds to the malicious behavior. The goal of the backdoor attack is to construct a poisoned dataset such that the model trained on it learns well for both the original normal distribution $\mathcal{D}_A$ and the malicious distribution $\mathcal{D}_B$. In other words, an attacker wants to construct a dataset $\tilde{D}$ such that the model trained on $\tilde{D}$, denoted $f_{\tilde{D}}$, has sufficiently low risk $\E_{(x_a, y_a)\sim\mathcal{D}_A} \ell(f_{\tilde{D}}, (x_a, y_a))$ and $\E_{(x_b, y_b)\sim\mathcal{D}_B} \ell(f_{\tilde{D}}, (x_b, y_b))$ at the same time.

One approach to constructing such a dataset $\tilde{D}$ is to directly mix the \textit{benign dataset} $D_A\sim\mathcal{D}_A^{N_A}$ and the \textit{trigger dataset} $D_B\sim\mathcal{D}_B^{N_B}$. An attacker usually wants to make the attack stealthy, and so it sets $N_B\ll N_A$. We define $\mathcal{D}_B$ according to the original normal behavior $\mathcal{D}_A$, the trigger $T\in\R^d$, and the trigger label $y_T\in\mathcal{Y}$:
\begin{equation}
    (x_b, y_b) := ((1-m)\odot x_a + m\odot T, y_T)\label{def:trig},
\end{equation}
where $x_a \sim \mathcal{D}_A$, $m\in\R^d$ is the real-valued mask, and $\odot$ is the Hadamard product. 

\section{Proposed Methods and Theoretical Analysis}\label{sec:3}

In this paper, we aim to use dataset distillation (KIP as a representative) to perform the backdoor attack. In the simplest form of KIP-based backdoor attacks (as shown in Algorithm~\ref{alg:bdKIP} of the Appendix), we first construct the \textit{poisoned dataset} $\tilde{D}=D_A \cup D_B$ from $\mathcal{D}_A^{N_A}$ and $\mathcal{D}_B^{N_B}$. Then, we perform KIP on $\tilde{D}$ and compress the information in $\tilde{D}$ into the \textit{distilled poisoned dataset} $\mathcal{S}^* = \{(x_s, y_s)\}_{s=1}^{N_\mathcal{S}}$, where $N_\mathcal{S} \ll N_A + N_B$. Namely, we solve the following optimization problem
\begin{equation}
    \mathcal{S}^* = \argmin_{\mathcal{S}} \frac{1}{N_A + N_B}\sum_{(x, y)\in \tilde{D}} \|f_\mathcal{S}(x) - y\|^2_2 \label{bdKIP:for},
\end{equation}
where $f_\mathcal{S}(x)^T = k(x, \mX_{\mathcal{S}}) [k(\mX_{\mathcal{S}}, \mX_{\mathcal{S}}) + N_{\mathcal{S}}\lambda\mI]^{-1} \mY_{\mathcal{S}}$. Essentially, the above KIP-based backdoor attack is the same as Naive attack in \citep{liu2023backdoor} except that the other distillation, instead of KIP, is used in Naive attack. The experimental results in \citep{liu2023backdoor} show that ASR grows but CTA drops significantly when the trigger size increases. \citet{liu2023backdoor} claims a trade-off between CTA and the trigger size. Nonetheless, we find that our KIP-based backdoor attack does not have such a trade-off. This motivates us to develop a theoretical framework for backdoor attacks on dataset distillation. 

Below, we introduce the theoretical framework in Section~\ref{sec: Theoretical Framework}, followed by two theory-induced backdoor attacks, \textsf{simple-trigger} and \textsf{relax-trigger} in Section~\ref{sec: Theory-Induced Backdoor: simple-trigger} and Section~\ref{sec: Theory-Induced Backdoor: relax-trigger}, respectively.

\subsection{Theoretical Framework}\label{sec: Theoretical Framework}
We first introduce the structure of our analysis, which divides the risk of KIP-based backdoor attacks into three parts: projection loss, conflict loss, and generalization gap. Then, we provide an upper bound for each part of the risk. 

\paragraph{Structure of Analysis.}
Recall that the goal of a KIP-based backdoor attack is to construct the synthetic dataset $\mathcal{S}^*$ such that the risk $\E_{(x, y) \sim\mathcal{D}}\ell(f_{\mathcal{S}^*}, (x, y))$ is sufficiently low, where $\mathcal{D}$ is the normal distribution $\mathcal{D}_A$ or the malicious distribution $\mathcal{D}_B$.
The classical framework for analyzing this problem is to divide the risk into two parts, the empirical risk and generalization gap. Namely, 
\begin{align}
    \E_{(x, y)\sim\mathcal{D}}~\ell(f_{\mathcal{S}^*}, (x, y)) 
    &= 
    \underbrace{\E_{(x, y)\sim D}~\ell(f_{\mathcal{S}^*}, (x, y))}_{\text{Empirical risk}}\nonumber\\
    &+ 
    \underbrace{[\E_{(x, y)\sim\mathcal{D}}~\ell(f_{\mathcal{S}^*}, (x, y)) - \E_{(x, y) \sim D}~\ell(f_{\mathcal{S}^*}, (x, y))]}_{\text{Generalization gap}}
\end{align}
where $D = \{(x_i, y_i)\}_{i=1}^N$ is the dataset sampled from the distribution $\mathcal{D}^N$ and $N$ is the number of samples of $D$. Here, we consider that $D$ is $D_A \sim\mathcal{D}^{N_A}$ or $D_B \sim\mathcal{D}^{N_B}$.  In our framework, we continue to divide the empirical risk into two parts as 
\begin{align}
    \E_{(x, y)\sim D}~\ell(f_{\mathcal{S}^*}, (x, y))
    &\leq
    \frac{N_A + N_B}{N}[
    \underbrace{\min_{\mathcal{S}} \E_{(x, y)\sim \tilde{D}}\ell(f_{\mathcal{S}}, (x, f_{\tilde{D}}(x)))}_{\text{Projection Loss}}
    + \underbrace{\E_{(x, y)\sim \tilde{D}}\ell(f_{\tilde{D}}, (x, y))}_{\text{Conflict Loss}}]
\end{align}
where $\tilde{D} = D_A  \cup D_B$, $f_{\tilde{D}}$ is the model trained on $\tilde{D}$ with the weight of the regularization term $\lambda\geq0$ and $f_{\mathcal{S}}$ is the model trained on $\mathcal{S}$ with the weight of the regularization term $\lambda_{\mathcal{S}}\geq 0$.
Intuitively, given a dataset $\tilde{D}$ constructed from $\mathcal{D}_A^{N_A}$ and $\mathcal{D}_B^{N_B}$, $f_{\tilde{D}}$ is regarded as the best model derived from the information of $\tilde{D}$. The conflict loss reflects the internal information conflict between the information about $\mathcal{D}_A$ in $\tilde{D}$ and the information about $\mathcal{D}_B$ in $\tilde{D}$.
{ For example, we consider a dog/cat picture classification problem. In the dataset $D_A$, we label the dog pictures with $0$ and label the cat pictures with $1$. However, in the dataset $D_B$, we label the dog pictures with $1$ and label the cat pictures with $0$. It is clear that the model trained on $\tilde{D}$ must perform terribly on the dataset either $D_A$ or $D_B$. In this case, the information between $D_A$ and $D_B$ have strong conflict and the conflict loss would be large.} 
On the other hand, projection loss reflects the loss of information caused by projecting $f_{\tilde{D}}$ into $\{f_\mathcal{S}|\mathcal{S} = \{(x_i, y_i)\in\mathcal{X}\times\mathcal{Y}\}_{i=1}^{N_{\mathcal{S}}}\}$. We can also consider the projection loss as the increase in information induced by compressing the information of $\tilde{D}$ into the synthetic dataset $\mathcal{S}$.
{ Take writing an abstract for example. If we want to write a 100 words abstract to  describe a 10000 words article, the abstract may suffer some lack of semantics to some degree. Such a phenomena also happens for dataset distillation. When the information of a large dataset is complex enough, the information loss for dataset distillation will be significant; When the information of a large dataset is very simple, it is possible that there is only very limited information loss. We introduce the projection loss defined above to measure this phenomenon.} 
More details can be found below.  

\paragraph{Conflict Loss.}
In a KIP-based backdoor attack, the dataset $\tilde{D}$ is defined as $\tilde{D} = D_A \cup D_B$, where $D_A \sim\mathcal{D}_A^{N_A}$ and $D_B\sim\mathcal{D}_B^{N_B}$. By Eq.~(\ref{sol:c-closed-form}) we know that the model trained on $\tilde{D}$ with the weight of the regularization term $\lambda\geq0$ has a closed-form solution if we constrain the model in the RKHS $\mathcal{H}_k^c$ and suppose that $\ell(f, (x, y)) := \|f(x) - y\|^2_2$:
\begin{equation}\label{sol:AB}
    f_{\tilde{D}}(x)^T = k(x, \mX_{AB}) [k(\mX_{AB}, \mX_{AB}) + (N_A + N_B)\lambda\mI]^{-1} \mY_{AB},
\end{equation}
where $(N_A + N_B) \times d$ matrix $\mX_{AB}$ is the matrix corresponding to the features of $\tilde{D}$, $(N_A + N_B) \times c$ matrix $\mY_{AB}$ is the matrix corresponding to the labels of $\tilde{D}$, $k(x, \mX_{\mathcal{AB}})$ is a $1\times (N_A + N_B)$ matrix, $k(\mX_{AB}, \mX_{AB})$ is a $(N_A + N_B)\times (N_A + N_B)$ matrix with $[k(\mX_{AB}, \mX_{AB})]_{ij} = k(x_{i}, x_{j})$, and $\mY_{AB}$ is a $(N_A + N_B)\times c$ matrix with $\mY_{AB} = [y_{1}, y_{2}, \dots, y_{(N_A + N_B)}]^T$.
Hence, we can express the conflict loss $\mathcal{L}_{\text{conflict}}$ as
\begin{align}
    \mathcal{L}_{\text{conflict}} 
    &=
    \frac{1}{N_A + N_B}
    \|\mY_{AB} - k(\mX_{AB}, \mX_{AB}) [k(\mX_{AB}, \mX_{AB}) + (N_A + N_B)\lambda\mI]^{-1} \mY_{AB}\|_2^2.\label{preds:AB}
\end{align}
We can obtain the upper bound of $\mathcal{L}_{\text{conflict}}$ as Theorem~\ref{thm:conflict}.

\begin{theorem}[Upper bound of conflict loss]\label{thm:conflict}
The conflict loss $\mathcal{L}_{\text{conflict}}$ can be bounded as
\begin{small}
\begin{align}
    \mathcal{L}_{\text{conflict}} 
    \leq
    \frac{1}{N_A + N_B}\text{Tr}(\mI - k(\mX_{AB}, \mX_{AB}) [k(\mX_{AB}, \mX_{AB}) + (N_A + N_B)\lambda\mI]^{-1})^2 \|\mY_{AB}\|^2_2
\end{align}
\end{small}
where $\text{Tr}$ is the trace operator, $k(\mX_{AB}, \mX_{AB})$ is a $(N_A + N_B)\times (N_A + N_B)$ matrix, and $\mY_{AB}$ is a $(N_A + N_B)\times c$ matrix. 
\end{theorem}

The proof of Theorem~\ref{thm:conflict} can be found in Appendix~\ref{sec: Theorem 1 and Its Proof}. From Theorem~\ref{thm:conflict}, we know that the conflict loss can be characterized by $\text{Tr}(\mI - k(\mX_{AB}, \mX_{AB}) [k(\mX_{AB}, \mX_{AB}) + (N_A + N_B)\lambda\mI]^{-1})$. However, in the latter sections, we do not utilize $\text{Tr}(\mI - k(\mX_{AB}, \mX_{AB}) [k(\mX_{AB}, \mX_{AB}) + (N_A + N_B)\lambda\mI]^{-1})$ to construct trigger pattern generalization algorithm; instead, we use Eq.~(\ref{preds:AB}) directly. It is because Eq.~(\ref{preds:AB}) can be computed more precisely although $\text{Tr}(\mI - k(\mX_{AB}, \mX_{AB}) [k(\mX_{AB}, \mX_{AB}) + (N_A + N_B)\lambda\mI]^{-1})$ and Eq.~(\ref{preds:AB}) have similar computational cost.

\paragraph{Projection Loss.}
To derive the upper bound of the projection loss, we first derive Lemma~\ref{lemma:proj}.

\begin{lemma}[Projection lemma]\label{lemma:proj}
Given a synthetic dataset $\mathcal{S} = \{(x_s, y_s)\}_{s=1}^{N_{\mathcal{S}}}$, and a dataset $\tilde{D} = \{(x_i, y_i)\}_{i=1}^{N_A + N_B}$ where $(N_A + N_B)$ is the number of the samples of $\tilde{D}$. Suppose the kernel matrix $k(\mX_{\mathcal{S}}, \mX_{\mathcal{S}})$ is invertible, then we have
\begin{scriptsize}
\begin{align}\label{proj:in}
    k(\cdot, x_i) 
    &= 
    \underbrace{k(\cdot, \mX_{\mathcal{S}})k(\mX_{\mathcal{S}}, \mX_{\mathcal{S}})^{-1}k(\mX_{\mathcal{S}}, x_i)}_{\in\mathcal{H}_\mathcal{S}}
    +
    \underbrace{[k(\cdot, x_i) -
    k(\cdot, \mX_{\mathcal{S}})k(\mX_{\mathcal{S}}, \mX_{\mathcal{S}})^{-1}k(\mX_{\mathcal{S}}, x_i)]}_{\in\mathcal{H}_{\mathcal{S}}^\perp}
    , \quad\forall(x_i, y_i) \in \tilde{D}
\end{align}
\end{scriptsize}

where $\mathcal{H}_{\mathcal{S}} := \text{span}(\{k(\cdot, x_s)\in\mathcal{H}_k| (x_s, y_s) \in \mathcal{S}\})$ and $\mathcal{H}_{\mathcal{S}}^\perp$ is the collection of functions orthogonal to $\mathcal{H}_{\mathcal{S}}$ corresponding to the inner product $\langle\cdot, \cdot\rangle_{\mathcal{H}_k}$.
Thus, $k(\cdot, \mX_{\mathcal{S}})k(\mX_{\mathcal{S}}, \mX_{\mathcal{S}})^{-1}k(\mX_{\mathcal{S}}, x_i)$ is the solution of the optimization problem:
\begin{small}
\begin{align}
    \argmin_{f\in \mathcal{H}_S} \sum_{(x_s, y_s)\in\mathcal{S}}\|f(x_s) - k(x_s, x_i)\|^2_2 .\label{sol:syn}
\end{align}
\end{small}
\end{lemma}

The proof of Lemma~\ref{lemma:proj} can be found in Appendix~\ref{sec: Lemma 1 and Its Proof}. Now, we turn to the scenario of the KIP-based backdoor attack. Given a mixed dataset $\tilde{D} = D_A \cup D_B$ where $D_A \sim\mathcal{D}^{N_A}_A$ and  $D_B \sim\mathcal{D}^{N_B}_B$. We also constrained models in the RKHS $\mathcal{H}_k^c$ and suppose $\ell(f, (x, y)) := \|f(x) - y\|^2_2$. With the help of Lemma~\ref{lemma:proj}, we can obtain the following theorem:

\begin{theorem}[Upper bound of projection loss]\label{thm:projloss}
Suppose the kernel matrix of the synthetic dataset $k(\mX_\mathcal{S}, \mX_\mathcal{S})$ is invertible, $f_{\mathcal{S}}$ is the model trained on the synthetic dataset $\mathcal{S}$ with the regularization term $\lambda_{\mathcal{S}}$, where the projection loss $\mathcal{L}_{\text{project}} = \min_{S} \E_{(x, y)\sim \tilde{D}}\ell(f_S, (x, f_{\tilde{D}}(x)))$ can be bounded as
\begin{small}
\begin{align}
    \mathcal{L}_{\text{project}} 
    \leq
    \sum_{(x_i, y_i)\in \tilde{D}}
    \min_{\mX_{\mathcal{S}}}
    \sum_{j=1}^c
    \frac{|\alpha_{i, j}|^2}{N_A + N_B}
    \|k(\mX_{AB}, x_i) - k(\mX_{AB}, \mX_\mathcal{S}) 
    k(\mX_\mathcal{S}, \mX_\mathcal{S})^{-1} k(\mX_\mathcal{S}, x_i)\|^2_2.\label{bound:project}
\end{align}
\end{small}
where $\alpha_{i, j}: = [[k(\mX_{AB}, \mX_{AB}) + (N_A + N_B)\lambda\mI]^{-1} \mY_{AB}]_{i, j}$,  which is the weight of $k(\cdot, x_i)$ corresponding to $f_{\tilde{D}}^j$, $\mX_{AB}$ is the $(N_A + N_B) \times d$ matrix corresponding to the features of $\tilde{D}$, $\mX_{\mathcal{S}}$ is the $N_{\mathcal{S}} \times d$ matrix corresponding to the features of $\mathcal{S}$, $\mY_{AB}$ is the $(N_A + N_B) \times c$ matrix corresponding to the labels of $\tilde{D}$, $\mY_{\mathcal{S}}$ is the $N_{\mathcal{S}} \times c$ matrix corresponding to the labels of $\mathcal{S}$.
\end{theorem}

The proof of Theorem~\ref{thm:projloss} can be found in Appendix~\ref{sec: Theorem 2 and Its Proof}. In Theorem~\ref{thm:projloss}, we first characterize the natural information loss when compressing the information of $\tilde{D}$ into an arbitrary dataset $\mathcal{S}$, and then bound the information loss for the synthetic dataset $\mathcal{S}^*$ generated by dataset compression by taking the minimum. This formulation gives some insight into the construction of our trigger generation algorithm, which is discussed in the later section.

\paragraph{Generalization Gap.}
Finally, for the generalization gap, we follow the existing theoretical results (Theorem 3.3 in \citep{10.5555/2371238}), but modify them a bit. Let $\mathcal{G} = \{g : (x,y)\mapsto \|f(x)-y\|^2_2 | f\in\mathcal{H}_k^c\}$. Assume that the distribution $\mathcal{D}$ is distributed in a bounded region, and that $\mathcal{G}\subset C^1$ and the norm of the gradient of $g\in\mathcal{G}$ have a common non-trivial upper bound. Namely, $\|(x, y) - (x', y')\|_2 \leq \Gamma_{\mathcal{D}}$ for any sample which is picked from $\mathcal{D}$ and $\|\nabla g\|_2 \leq L_{\mathcal{D}}$. Then we can obtain Theorem~\ref{thm:gen}.
\begin{theorem}[Upper bound of generalization gap]\label{thm:gen}
Given a $N$-sample dataset $D$, sampled from the distribution $\mathcal{D}$, the following generalization gap holds for all $g\in\mathcal{G}$ with probability at least $1-\delta$:
\begin{small}
\begin{equation}
    \E_{(x,y)\sim\mathcal{D}}[g((x, y))] - \sum_{(x_i, y_i)\in D} \frac{g((x_i, y_i))}{N} 
    \leq 
    2\Hat{\mathfrak{R}}_{D}(\mathcal{G})
    + 3 L_{\mathcal{D}} \Gamma_{\mathcal{D}} \sqrt{\frac{\log{\frac{2}{\delta}}}{2N}}\label{eq:gengap},
\end{equation}
\end{small}

where $\mX$ is the matrix of the features of $D$ and $\Hat{\mathfrak{R}}_{D}(\mathcal{G})$ is the empirical Rademacher's complexity.
\end{theorem}
The proof of Theorem~\ref{thm:gen} can be found in Appendix~\ref{sec: Theorem 3 and Its Proof}. We know from Theorem~\ref{thm:gen} that the upper bound of the generalization gap is characterized by two factors, $\Hat{\mathfrak{R}}_{D}(\mathcal{G})$ and $\Gamma_{\mathcal{D}}$. The lower $\Hat{\mathfrak{R}}_{D}(\mathcal{G})$ and $\Gamma_{\mathcal{D}}$ imply the lower generalization gap. We usually assume $k(x, x) \leq r^2$ and $\sqrt{\langle f, f\rangle_{\mathcal{H}_k}}\leq\Lambda$ (as in Theorem 6.12 of \citep{10.5555/2371238}). Under this setting, we can ignore $\Hat{\mathfrak{R}}_{D}(\mathcal{G})$ for the upper bound of the generalization gap and only focus on $\Gamma_{\mathcal{D}}$. ASR relates to the risk for $\mathcal{D}_B$ and hence corresponds to the generalization gap evaluated on $\mathcal{D}_B$. This theoretical consequence can be used to explain the phenomenon that ASR of the backdoor attack increases as we enlarge the trigger size. 

\subsection{Theory-Induced Backdoor: \textsf{simple-trigger}}\label{sec: Theory-Induced Backdoor: simple-trigger}

Consider $\mathcal{D}$ in Theorem~\ref{thm:gen} as $\mathcal{D}_B$ and the corresponding dataset $D$ as $D_B$. Conventionally, a cell of the mask $m$ in Eq.~(\ref{def:trig}) is $1$ it corresponds to a trigger, and is 0 otherwise. Recall that the definition of $\mathcal{D}_B$ in Eq.~(\ref{def:trig}), it is clear that the $\Gamma_{\mathcal{D}_B}$ will monotonely decrease from $\Gamma_{\mathcal{D}_A}$ to $0$ as we enlarge the trigger size. If we enlarge the trigger size, the $\Gamma_{\mathcal{D}_B}$ drops to zero, which implies that the corresponding generalization gap will be considerably small. Thus, the success of the large trigger pattern can be attributed to its relatively small generalization gap.

So, given an image of size $m\times n$ ($m\leq n$), \textsf{simple-trigger} generates a trigger of size $m\times n$. The default pattern for the trigger generated by  \textsf{simple-trigger} is whole-white. In fact, since the generalization gap is irrelevant to the trigger pattern, we do not impost any pattern restrictions. 

\subsection{Theory-Induced Backdoor: \textsf{relax-trigger}}\label{sec: Theory-Induced Backdoor: relax-trigger}

In \textsf{simple-trigger}, we optimize the trigger through only the generalization gap. However, we know that ASR can be determined by conflict loss, projection loss, and generalization gap because of Theorems~\ref{thm:conflict}$\sim$\ref{thm:gen} (i.e., all are related to $\mathcal{D}_B$). On the other hand, CTA is related to conflict loss and projection loss, because the generalization gap is irrelevant to CTA. That is, Eq.~(\ref{eq:gengap}) evaluated on $\mathcal{D}_A$ is a constant as we modify the trigger. As a result, the lower conflict loss, projection loss, and generalization gap imply a backdoor attack with greater ASR and CTA. Therefore, \textsf{relax-trigger} aims to construct a trigger whose corresponding $\mathcal{D}_B$ make Eq.~(\ref{preds:AB}), Eq.~(\ref{bound:project}), and $\Gamma_{\mathcal{D}_B}$ sufficiently low. The computation procedures of \textsf{relax-trigger} can be found in Algorithm~\ref{alg:triggen} of Appendix~\ref{sec: Pseudocode for relax-trigger}. 

Suppose $D_A$, $N_A$ and $N_B$ are fixed. To reduce the bound in Eq.~(\ref{preds:AB}), one considers $D_B$ as a function depending on the trigger $T$ and then uses the optimizer to find the optimal trigger $T^*$. In this sense, we solve the following optimization problem
\begin{small}
\begin{equation}
        \argmin_{T}\|\mY_{AB} - k(\mX_{AB}, \mX_{AB}) [k(\mX_{AB}, \mX_{AB}) + (N_A + N_B)\lambda\mI]^{-1} \mY_{AB}\|_2^2.
\end{equation}
\end{small}

On the other hand, a low $\Gamma_{\mathcal{D}_B}$ can be realized by enlarging the trigger as mentioned in Section~\ref{sec: Theory-Induced Backdoor: simple-trigger}.

Finally, to make Eq.~(\ref{bound:project}) sufficiently low, we consider $\mathcal{D}_B$ as a function of the trigger $T$, and then directly optimize
\begin{small}
\begin{equation}\label{alg:1}
    \argmin_{T}
    \left\{
    \sum_{(x_i, y_i)\in \tilde{D}}
    \min_{\mX_{\mathcal{S}}}
    \sum_{j=1}^c
    |\alpha_{i, j}|^2
    \|k(\mX_{AB}, x_i) - k(\mX_{AB}, \mX_\mathcal{S}) 
    k(\mX_\mathcal{S}, \mX_\mathcal{S})^{-1} k(\mX_\mathcal{S}, x_i)\|^2_2
    \right\}.
\end{equation}
\end{small}

However, Eq.~(\ref{alg:1}) is a bi-level optimization problem that is difficult to solve. Instead, we set the synthetic dataset $\mathcal{S}$ in Eq.~(\ref{alg:1}) to $\mathcal{S}_A$, which is the distilled dataset from $D_A$. Then, the two-level optimization problem can be converted into a one-level optimization problem below.
\begin{equation}
    \argmin_{T} 
    \left\{
    \sum_{(x_i, y_i)\in \tilde{D}}
    \sum_{j=1}^c
    |\alpha_{i,j}|^2
    \|k(\mX_{AB}, x_i) - k(\mX_{AB}, \mX_{\mathcal{S}_A}) 
    k(\mX_{\mathcal{S}_A}, \mX_{\mathcal{S}_A})^{-1} k(\mX_{\mathcal{S}_A}, x_i)\|^2_2
    \right\}.\label{alg:2}
\end{equation}
Eq.~(\ref{alg:2}) can be easily solved by directly applying optimizers like Adam~\citep{adam}. Eq.~(\ref{alg:2}) aims to find a trigger $T$ such that $\tilde{D}$ generated from $D_A$ and $D_B$ will be compressed into the neighborhood of $\mathcal{S}_A$ $\subset (\mathcal{X}\times \mathcal{Y})^{N_\mathcal{S}}$, which guarantees that CTA of the model trained on the distilled $\tilde{D}$ is similar to CTA of the model trained on the distilled $D_A$. Overall, \textsf{relax-trigger} solves the following optimization,
\begin{align}
    &\argmin_{T} 
    \{
    \sum_{(x_i, y_i)\in \tilde{D}}
    \sum_{j=1}^c
    |\alpha_{i, j}|^2
    \|k(\mX_{AB}, x_i) - k(\mX_{AB}, \mX_{\mathcal{S}_A}) 
    k(\mX_{\mathcal{S}_A}, \mX_{\mathcal{S}_A})^{-1} k(\mX_{\mathcal{S}_A}, x_i)\|^2_2\nonumber\\
    &+
    \rho\|\mY_{AB} - k(\mX_{AB}, \mX_{AB}) [k(\mX_{AB}, \mX_{AB}) + (N_A + N_B)\lambda\mI]^{-1} \mY_{AB}\|_2^2\}\label{eq:triggen},
\end{align}
where $\rho>0$ is the penalty parameter, $m$ is the previously chosen mask, the malicious dataset is defined as $D_B = \{(x_b, y_b) = ((1-m)\odot x_a + m\odot T, y_T)| (x_a, y_a)\in D_A\}$. We particularly note that Eq.~(\ref{alg:1}) is converted into Eq.~(\ref{alg:2}) because we use $\mathcal{S}_A$ to replace the minimization over $\mathcal{S}$. 

\textsf{relax-trigger} is different from DOORPING in \citep{liu2023backdoor}. DOORPING generates the trigger during the process of sample compression. In other words, DOORPING is induced by solving a bi-level optimization problem. However, \textsf{relax-trigger} is induced by a one-level optimization problem (Eq.~(\ref{eq:triggen})). The design rationale  of \textsf{relax-trigger} is different from DOORPING. DOORPING aims to find the globally best trigger but consumes a significant amount of computation time. On the other hand, through our theoretical framework, \textsf{relax-trigger} aims to find the trigger that reliably compresses the corresponding $\tilde{D}$ into the neighborhood of our $\mathcal{S}_A$ with the benefit of time efficiency.

\section{Evaluation}\label{sec: Experiment}

\subsection{Experimental Setting}\label{sec: Experimental Setting}
\paragraph{Dataset.} Two datasets are chosen for measuring the backdoor performance.
\begin{itemize}[leftmargin=*]
    \item \textbf{CIFAR-10} is a 10-class dataset with $6000$ $32\times32$ color images per class. CIFAR-10 is split into $50000$ training images and $10000$ testing images.

    \item \textbf{GTSRB} contains 43 classes of traffic signs with $39270$ images, which are split into $26640$ training images and $12630$ testing images. We resize all images to $32\times32$ color images.
\end{itemize}

\paragraph{Dataset Distillation and Backdoor Attack.} We use KIP \citep{nguyen2020dataset} to implement backdoor attacks with the neural tangent kernel (NTK) induced by a 3-layer neural network, which has the same structure in the Colab notebook of \citep{nguyen2020dataset}. We also set the optimizer to Adam~\citep{adam}, the learning rate to $0.01$, and the batch size to $10\times\text{number of class}$ for each dataset. We run KIP with $1000$ training steps to generate a distilled dataset. We perform 3 independent runs for each KIP-based backdoor attack to examine the performance.

\paragraph{Evaluation Metrics.} We consider two metrics, clean test accuracy (CTA) and attack success rate (ASR). Consider $\mathcal{S}$ as a distilled dataset from the KIP-based backdoor attack. CTA is defined as the test accuracy of the model trained on $\mathcal{S}$ and evaluated on the normal (clean) test dataset, while ASR is defined as the test accuracy of the model trained on $\mathcal{S}$ and evaluated on the trigger test dataset.

\textbf{Defense for Backdoor Attack.} In this paper we consider eight existing defenses, SCAn \citep{tang2021demon}, AC \citep{chen2018detecting}, SS \citep{tran2018spectral}, Strip (modified as a poison cleaner) \citep{gao2019strip}, ABL \citep{li2021anti}, NAD \citep{li2021neural}, STRIP (backdoor input filter) \citep{gao2019strip}, FP \citep{liu2018fine}, to investigate the ability to defend against KIP-based backdoor attack. The implementation of the above defenses is from the backdoor-toolbox\footnote{Available at \url{https://github.com/vtu81/backdoor-toolbox}.}. 

\subsection{Experimental Results}\label{sec: Experimental Results}
\paragraph{Performance of \textsf{simple-trigger}.} 
We performed a series of experiments to demonstrate the effectiveness of \textsf{simple-trigger}. In our setting, $N_{\mathcal{S}}$ is set to $10\times\text{number of classes}$ and $50\times\text{number of classes}$ for each dataset. We also configurated the trigger as $2\times2$, $4\times4$, $8\times8$, $16\times16$, $32\times32$ white square patterns. The corresponding results are shown in Table~\ref{tab:per_simple}. 
The experiment results suggest that CTA and ASR of \textsf{simple-trigger} increase as we enlarge the trigger size, which is consistent with our theoretical analysis (Theorem~\ref{thm:gen}). One can see that for the $32\times32$ white square trigger, ASR can achieve $100\%$ without sacrificing CTA.

\begin{table}[ht]
    \centering
    \resizebox{\textwidth}{!}{
    \begin{tabular}{|c||c|c|c|c|c|c|c|c|c|c|c|c|}
         \hline   
         $\text{Data. (Size)}\backslash\text{Trig.}$ & None & \multicolumn{2}{|c|}{$2\times2$} & \multicolumn{2}{|c|}{$4\times4$} & \multicolumn{2}{|c|}{$8\times8$} & \multicolumn{2}{|c|}{$16\times16$} & \multicolumn{2}{|c|}{$32\times32$}\\
         \hline
         \multicolumn{1}{|c||}{} & \multicolumn{1}{|c}{CTA (\%)}  & \multicolumn{1}{|c}{CTA (\%)} & \multicolumn{1}{c|}{ASR (\%)} & \multicolumn{1}{|c}{CTA (\%)} & \multicolumn{1}{c|}{ASR (\%)} & \multicolumn{1}{|c}{CTA (\%)} & \multicolumn{1}{c|}{ASR (\%)} & \multicolumn{1}{|c}{CTA (\%)} & \multicolumn{1}{c|}{ASR (\%)} & \multicolumn{1}{|c}{CTA (\%)} & \multicolumn{1}{c|}{ASR (\%)}\\
         \hline
         CIFAR-10 (100) & 42.55 (0.13) & 41.78 (0.22) & 65.73 (0.80) & 41.53 (0.31) & 90.59 (0.22)	& 41.46 (0.32) & 98.29 (0.18) & 41.55 (0.43) & 99.94 (0.05) & 41.70 (0.25) & 100.00 (0.00)\\
         \hline
         CIFAR-10 (500) & 44.52 (0.23) & 43.89 (0.13) & 82.36 (0.39) & 43.85 (0.23) & 92.89 (0.16) & 43.60 (0.23) & 98.19 (0.16) & 43.70 (0.40) & 99.88 (0.07) & 43.66 (0.40) & 100.00 (0.00)\\														
         \hline
         GTSRB (430) & 69.27 (0.19) & 67.06 (0.74) & 74.14 (0.50) & 67.01 (0.69) & 81.46 (0.37) & 66.98 (0.64) & 89.63 (0.58) & 67.10 (0.63) & 98.43 (0.13) & 67.56 (0.60) &	100.00 (0.00)\\
         \hline
         GTSRB (2150) & 72.07 (0.20) & 70.87 (0.27) & 76.79 (1.08) & 70.90 (0.25) & 81.93 (0.62) & 70.92 (0.31) & 90.48 (0.74) & 70.98 (0.22) & 98.89 (0.17) & 71.27 (0.24) & 100.00 (0.00)\\
         \hline
    \end{tabular}}
    \caption{Performance of \textsf{simple-trigger} on CIFAR-10 and GTSRB (mean and standard deviation).}
    \label{tab:per_simple}
        \vspace{-4mm}
\end{table}

\paragraph{Performance of \textsf{relax-trigger}.} Here, we relax the setting of the mask $m$; i.e., each component of $m$ is defined to be $0.3$, instead of $1$. This can be regarded as an increase in the trigger's transparency (the level of invisibility) for mixing an image and the trigger. Recall the definition of $\mathcal{D}_B$ in (Eq. \ref{def:trig}). From theory point of view, under such a mask $m$, $\Gamma_{\mathcal{D}_B}$ will drop to $0.3 *\Gamma_{\mathcal{D}_A} > 0$, as we enlarge the trigger. Hence, we cannot reduce the generalization gap considerably as in the experiments of \textsf{simple-trigger}. It turns out that to derive better CTA and ASR, we resort to consider \textsf{relax-trigger}.

\begin{figure}
    \centering
    \subcaptionbox{{\tiny CIFAR10 (size 100)}}{
        \includegraphics[width = .14\linewidth]{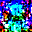}
    }\hfill
    \subcaptionbox{{\tiny CIFAR10 (size 500)}}{
        \includegraphics[width = .14\linewidth]{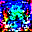}
    }\hfill
    \subcaptionbox{{\tiny GTSRB (size 430)}}{
        \includegraphics[width = .14\linewidth]{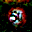}
    }\hfill    \textbf{\subcaptionbox{{\tiny GTSRB (size 2150)}}{
        \includegraphics[width = .14\linewidth]{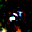}
    }}
    \caption{Triggers generated by \textsf{relax-trigger} for GTSRB and CIFAR.}
    \label{fig:relax-trig}
    \vspace{-4mm}
\end{figure}

The result is presented in Table~\ref{tab:per_relax}. We compare the performance (CTA and ASR) between \textsf{simple-trigger} ($32 \times 32$ white square), DOORPING  and \textsf{relax-trigger}. For CIFAR-10,  \textsf{relax-trigger} increases the ASR about $24\%$ from \textsf{simple-trigger} without losing CTA. For GTSRB,  \textsf{relax-trigger} not only increases the ASR about $30\%$, but also slightly increases the CTA. On the other hand, \textsf{relax-trigger} possesses higher CTA and ASR compared to DOORPING. These results confirm the effectiveness of \textsf{relax-trigger}. The trigger patterns of \textsf{relax-trigger} are visualized in Figure \ref{fig:relax-trig}. 

\begin{table}[ht]
    \centering
    \scalebox{0.6}[0.6]{
    \begin{tabular}{|c|c||c|c|c|c|c|c|}
         \hline   
         Dataset & $\text{Size}\backslash\text{Trig.}$ & \multicolumn{2}{|c|}{\textsf{simple-trigger} (baseline)} & \multicolumn{2}{|c|}{\textsf{relax-trigger}} & \multicolumn{2}{|c|}{DOORPING}\\
         \hline
         \multicolumn{2}{|c||}{} & \multicolumn{1}{|c}{CTA (\%)} & \multicolumn{1}{c|}{ASR (\%)} & \multicolumn{1}{|c}{CTA (\%)} & \multicolumn{1}{c|}{ASR (\%)} & \multicolumn{1}{|c}{CTA (\%)} & \multicolumn{1}{c|}{ASR (\%)}\\
         \hline
         CIFAR-10 & 100 &  41.40 (0.06) & 75.92 (1.19) & 41.66 (0.74) & 100.00 (0.00) & 36.35 (0.42) & 80.00 (40.00)\\
         \hline
         CIFAR-10 & 500 & 42.98 (0.13) & 75.79 (0.58) & 43.64 (0.40) & 100.00 (0.00) &  & \\
         \hline
         GTSRB & 430 &  67.02 (0.07) & 62.74 (0.23) & 68.73 (0.67) & 95.26 (0.54) & 68.03 (0.92) & 90.00 (30.00)\\
         \hline
         GTSRB & 2150 & 70.28 (0.07) & 62.65 (1.12) & 71.54 (0.33)	 & 95.08 (0.33) &  &\\
         \hline
    \end{tabular}}
    \caption{Performance of \textsf{relax-trigger} on CIFAR-10 and GTSRB (mean and standard deviation).}
    \label{tab:per_relax}
        \vspace{-4mm}
\end{table}

\paragraph{Off-the-shelf Backdoor Defenses.}
We examine whether \textsf{simple-trigger} and \textsf{relax-trigger} can survive backdoor detection and cleansing. Here, we utilize  backdoor-toolbox and retrain the distilled dataset on ResNet (default setting in backdoor-toolbox) to compute CTA and ASR. In our experimental results, the term ``None'' denotes no defense. 

For \textsf{simple-trigger}, we find that both CTA and ASR of None increase as we enlarge the trigger size. Moreover, both CTA and ASR of None increase as we enlarge the size of the distilled dataset. The above implies that \textsf{simple-trigger} is more suitable for large-size distilled datasets. Since the CTA and ASR increase as we enlarge the trigger, we focus on $32\times 32$ trigger images in the following discussion. In the case of CIFAR-10, for size 100 (see Table~\ref{tab:def:cifar10-100}), we can find that ASR of NAD is still $1$. That is, NAD fails to remove the backdoor. For the other defenses, the CTA drops over $7\%$, though they can reduce the ASR. Hence, we conclude that these defenses are not effective. For size 500 (see Table~\ref{tab:def:cifar10-500} in Appendix~\ref{sec: Extra Experiments}), the ASR of SCAn is still $1$, implying that SCAn fails to remove the backdoor. 
The other defenses, SS, Strip, ABL, STRIP, and FP considerably compromise the CTA. Overall, the above results also suggest that the defenses may be more successful when we increase the size of the distilled dataset. On the other hand, for GTSRB (see Tabel~\ref{tab:def:gtsrb-430} and Table~\ref{tab:def:gtsrb-2150} in Appendix~\ref{sec: Extra Experiments}), we also reach a similar conclusion.

For \textsf{relax-trigger} (see Table~\ref{tab:def:relax-trig} in Appendix~\ref{sec: Extra Experiments}), all defenses considered in this paper cannot effectively remove the backdoor. In particular, in the case of CIFAR-10, for size 100, SCAn, AC, Strip, and ABL do not reduce the ASR. They even increase ASR to some degree. On the other hand, SS, STRIP, and FP also compromise the CTA too much. Lastly, though NAD reaches a better defense result; however, the corresponding ASR still remains about $50\%$ of None's ASR. Essentially, this suggests that NAD cannot completely defend against \textsf{relax-trigger}. 
For the other defenses, the ASR still remains over $30\%$ of None's ASR. These defenses are ineffective against \textsf{relax-trigger}.

In the case of GTSRB, for size 430, we can also find that SCAn, NAD, and STRIP cannot successfully remove the backdoor. The ASR still remains over $70\%$ of None's ASR. Besides, we can find that AC, SS, Stip, ABL, and FP still compromise the CTA too much. Finally, for size 2150, AC, Strip, NAD, and STRIP still remain ASR over $50\%$ of None's ASR. Furthermore, SCAn, ABL, and FP even increase the ASR. In addition, SS decreases the CTA by about $45\%$ of None's CTA. To sum up, \textsf{relax-trigger} shows strong backdoor resiliency against all the tested defenses. 
\begin{table}[t]
    \centering
    \resizebox{\textwidth}{!}{
    \begin{tabular}{|c||c|c|c|c|c|c|c|c|c|c|c|c|c|c|c|c|c|}
    \hline
     $\text{Trig.}\backslash\text{Def.}$ & \multicolumn{2}{|c|}{None} & \multicolumn{2}{|c|}{SCAn} & \multicolumn{2}{|c|}{AC} & \multicolumn{2}{|c|}{SS} & \multicolumn{2}{|c|}{Strip}\\ 
    \hline
    \multicolumn{1}{|c||}{} & \multicolumn{1}{|c}{CTA (\%)} & \multicolumn{1}{c|}{ASR (\%)} & 
    \multicolumn{1}{|c}{CTA (\%)} & \multicolumn{1}{c|}{ASR (\%)} & \multicolumn{1}{|c}{CTA (\%)} & \multicolumn{1}{c|}{ASR (\%)} & \multicolumn{1}{|c}{CTA (\%)} & \multicolumn{1}{c|}{ASR (\%)} & \multicolumn{1}{|c}{CTA (\%)} & \multicolumn{1}{c|}{ASR (\%)}\\
    \hline
    $2\times 2$ & 23.18 (1.24) & 13.98 (8.36) & 24.39 (2.23) & 18.17 (2.74) & 23.84 (1.12) & 10.57 (3.52) & 22.60 (1.62) & 11.64 (1.42) & 25.08 (0.48) & 12.79 (5.25)\\
    \hline
    $4\times 4$ & 23.18 (1.40) & 25.26 (9.67) & 24.67 (0.97) & 15.73 (3.69) \% & 24.37 (1.29) & 14.00 (5.84) & 21.98 (3.30) &	10.81 (2.44) & 24.28 (0.39) & 17.68 (5.47)\\ 
    \hline
    $8\times 8$ & 25.69 (1.06) &	13.35 (5.38) & 26.40 (0.11) & 14.08 (3.72) & 23.24 (1.96) & 9.19 (5.53) & 21.49 (1.77) & 7.10 (4.61) & 25.13 (0.74) & 12.09 (5.52)\\ 
    \hline
    $16\times 16$ & 25.90 (5.76) & 81.29 (2.96) & 26.39 (2.96) & 49.66 (9.66) & 25.85 (1.57) & 55.26 (10.94) & 24.03 (1.66) & 40.03 (27.29) & 26.22 (0.75) & 41.36 (40.18)\\ 
    \hline
    $32\times 32$ & 28.95 (1.56) & 100.00 (0.00) & 28.28 (1.45) & 66.67 (47.14) & 25.35 (2.05) &	66.67 (47.14) & 22.21 (1.02) & 66.67 (47.14) & 25.68 (2.04) &	0.00 (0.00)\\ 
    \hline\hline
    $\text{Trig.}\backslash\text{Def.}$ 
    & \multicolumn{2}{|c|}{}& \multicolumn{2}{|c|}{ABL} & \multicolumn{2}{|c|}{NAD} & \multicolumn{2}{|c|}{STRIP} & \multicolumn{2}{|c|}{FP}\\
    \hline
    \multicolumn{1}{|c||}{} & \multicolumn{1}{|c}{} & \multicolumn{1}{c|}{} & 
    \multicolumn{1}{|c}{CTA (\%)} & \multicolumn{1}{c|}{ASR (\%)} & \multicolumn{1}{|c}{CTA (\%)} & \multicolumn{1}{c|}{ASR (\%)} & \multicolumn{1}{|c}{CTA (\%)} & \multicolumn{1}{c|}{ASR (\%)} & \multicolumn{1}{|c}{CTA (\%)} & \multicolumn{1}{c|}{ASR (\%)}\\
    \hline
    $2\times 2$ 
    & & & 13.31 (2.43) & 1.38 (1.14) & 31.74 (1.90) &	5.45 (0.78) & 20.91 (1.07) & 12.63 (7.61) & 13.05 (1.33) & 21.85 (30.90)\\
    \hline
    $4\times 4$ 
    & & & 13.12 (2.04) & 13.46 (13.00) & 30.87 (3.23) &	7.86 (4.36)	& 20.95 (1.15) & 19.08 (4.01) & 13.11 (1.43) & 73.25 (12.25)\\
    \hline
    $8\times 8$ 
    & & & 14.10 (0.47) & 24.92 (34.63) & 33.05 (1.04) & 10.82 (5.21) & 23.07 (1.01) & 11.84 (4.84) & 15.27 (1.77) & 2.81 (0.66)\\
    \hline
    $16\times 16$ 
    & & & 14.56 (2.67) & 35.47 (36.63) & 32.77 (1.66) & 22.25 (4.21) & 23.35 (0.30) & 66.53 (10.35) & 15.54 (0.21) &	22.94 (32.37)\\
    \hline
    $32\times 32$ 
    & & & 16.25 (4.23) & 33.33 (47.14) & 33.22 (3.78) & 100.00 (0.00) & 26.03 (1.33) & 0.00 (0.00) & 18.15 (1.38) & 0.00 (0.00) \\
    \hline
    \end{tabular}}
    \caption{Defenses for \textsf{simple-trigger} on CIFAR-10 with distilled dataset size = $100$.}
    \label{tab:def:cifar10-100}
        \vspace{-2mm}
\end{table}

\section{Conclusion}\vspace{-0.3cm}
In this paper, we present a novel theoretical framework based on the kernel inducing points (KIP) method to study the interplay between backdoor attacks and dataset distillation. The backdoor effect is characterized by three key components: conflict loss, projection loss, and generalization gap, along with two theory-induced attacks, \textsf{simple-trigger} and \textsf{relax-trigger}. Our \textsf{simple-trigger} proves that enlarged trigger size leads to improved ASR without sacrificing CTA.
Our \textsf{relax-trigger} presents a new and resilient backdoor attack scheme that either completely breaks or significantly weakens eight existing backdoor defense methods. Our study provides novel theoretical insights, unveils new risks of dataset distillation-based backdoor attacks, and calls for better defenses.

\clearpage
\bibliography{iclr2024_conference}
\bibliographystyle{iclr2024_conference}
\clearpage
\appendix
\section{Appendix}
\subsection{Notation Table}\label{sec: Notation Table}
The notations used in this paper are presented in Table~\ref{table:notations}.

\begin{table}[ht]
    \centering
    \begin{tabular}{p{1in}p{3.25in}}
        \multicolumn{1}{l}{Notations} & \multicolumn{1}{c}{Descriptions}\\
        \hline
        $\mathcal{X}$ & feature space $\subset\R^d$\\
        $\mathcal{Y}$ & label space $\subset\R^c$\\
        $x$ & feature\\
        $y$ & label\\
        $\mathcal{D}$ & probability distribution which is distributed in $\mathcal{X}\times\mathcal{Y}$\\
        $\mathcal{D}_A$ & probability distribution which is distributed in $\mathcal{X}\times\mathcal{Y}$ for benign behaviors\\
        $\mathcal{D}_B$ &  probability distribution which is distributed in $\mathcal{X}\times\mathcal{Y}$ for malicious behavior (trigger)\\
        $N$ & number of samples for some dataset\\
        $N_A$ & number of samples for benign dataset\\
        $N_B$ & number of samples for trigger dataset\\
        $N_{\mathcal{S}}$ & number of samples for distilled dataset\\
        $D$ & dataset picked from the distribution $\mathcal{D}$ with $N$ samples\\
        $D_A$ & dataset picked from the distribution $\mathcal{D}_A$ with $N_A$ samples\\
        $D_B$ & dataset picked from the distribution $\mathcal{D}_B$ with $N_B$ samples\\
        $\mathcal{S}$ &  any dataset with $N_{\mathcal{S}}$ samples\\
        $\mathcal{S}^*$ &  distilled dataset with $N_{\mathcal{S}}$ samples\\
        $\mathcal{S}^*_A$ &  distilled dataset from $D_A$ with $N_{\mathcal{S}}$ samples\\
        $T$ & trigger pattern $\in\R^d$\\
        $y_T$ & trigger label $\in\mathcal{Y}$\\
        $\tilde{D}$ & poisoned dataset which is the union from $D_A$ and $D_B$\\
        $\mX$ & the $N\times d$ matrix induced from the feature set in $D$.\\
        $\mY$ & the $N\times c$ matrix induced from the label set in $D$.\\
        $\mX_A$ & the $N_A\times d$ matrix induced from the feature set in $D_A$.\\
        $\mY_A$ & the $N_A\times c$ matrix induced from the label set in $D_A$.\\
        $\mX_B$ & the $N_B\times d$ matrix induced from the feature set in $D_B$.\\
        $\mY_B$ & the $N_B\times c$ matrix induced from the label set in $D_B$.\\
        $\mX_{\mathcal{S}}$ & the $N_{\mathcal{S}}\times d$ matrix induced from the feature set in $\mathcal{S}$.\\
        $\mY_{\mathcal{S}}$ & the $N_{\mathcal{S}}\times c$ matrix induced from the label set in $\mathcal{S}$.\\
        $\mX_{AB}$ & the $(N_{A} + N_{B})\times d$ matrix induced from the feature set in $\tilde{D}$.\\
        $\mY_{AB}$ & the $(N_{A} + N_{B})\times c$ matrix induced from the label set in $\tilde{D}$.\\
        $k(\cdot, \cdot)$ & the kernel \\
        $\mathcal{H}_k$ & the reproducing kernel hilbert space induced by kernel $k$.\\
        $\lambda$ & weight of regularization term.\\
        $\lambda_{\mathcal{S}}$ & weight of regularization term for $\mathcal{S}$.\\
        $\rho$ & penalty parameter.\\
        $f_{\tilde{D}}$  & the model trained on $\tilde{D}$ with the weight of the regularization term $\lambda\geq0$.\\ 
        $f_{\mathcal{S}}$ & the model trained on $\mathcal{S}$ with the weight of the regularization term $\lambda_{\mathcal{S}}\geq 0$.\\
        \hline
    \end{tabular}
    \caption{Notation Table}
    \label{table:notations}
\end{table}

\subsection{Lemma 1 and Its Proof}\label{sec: Lemma 1 and Its Proof}
\begin{customlemma}{1}[Projection lemma]\label{Apd:lemma:proj}
Given a synthetic dataset $\mathcal{S} = \{(x_s, y_s)\}_{s=1}^{N_{\mathcal{S}}}$, and a dataset $\tilde{D} = \{(x_i, y_i)\})_{i=1}^{N_A + N_B}$ where $(N_A + N_B)$ is the number of the samples of $\tilde{D}$. Suppose the kernel matrix $k(\mX_{\mathcal{S}}, \mX_{\mathcal{S}})$ is invertible, then we have
\begin{align}
    k(\cdot, x_i) 
    &= 
    \underbrace{k(\cdot, \mX_{\mathcal{S}})k(\mX_{\mathcal{S}}, \mX_{\mathcal{S}})^{-1}k(\mX_{\mathcal{S}}, x_i)}_{\in\mathcal{H}_\mathcal{S}}\label{Apd:proj:in}\\
    &+
    \underbrace{[k(\cdot, x_i) -
    k(\cdot, \mX_{\mathcal{S}})k(\mX_{\mathcal{S}}, \mX_{\mathcal{S}})^{-1}k(\mX_{\mathcal{S}}, x_i)]}_{\in\mathcal{H}_{\mathcal{S}}^\perp}
    , \quad\forall (x_i, y_i) \in \tilde{D}\label{Apd:proj:orth}
\end{align}
where $\mathcal{H}_{\mathcal{S}} := \text{span}(\{k(\cdot, x_s)\in\mathcal{H}_k| (x_s, y_s) \in \mathcal{S}\})$ and $\mathcal{H}_{\mathcal{S}}^\perp$ is the collection of functions which is orthogonal to $\mathcal{H}_{\mathcal{S}}$ corresponding to the inner product $\langle\cdot, \cdot\rangle_{\mathcal{H}_k}$.
The right hand side of (\ref{Apd:proj:in}) lies in $\mathcal{H}_{\mathcal{S}}$ while (\ref{Apd:proj:orth}) lies in $\mathcal{H}_{\mathcal{S}}^\perp$.
Thus, $k(\cdot, \mX_{\mathcal{S}})k(\mX_{\mathcal{S}}, \mX_{\mathcal{S}})^{-1}k(\mX_{\mathcal{S}}, x_i)$ is the solution of the optimization problem:
\begin{align}
    \argmin_{f\in \mathcal{H}_S} \sum_{(x_s, y_s)\in\mathcal{S}}\|f(x_s) - k(x_s, x_i)\|^2_2 .\label{Apd:sol:syn}
\end{align}
\end{customlemma}

\begin{proof} $k(\cdot, \mX_{\mathcal{S}})k(\mX_{\mathcal{S}}, \mX_{\mathcal{S}})^{-1}k(\mX_{\mathcal{S}}, x_i)$ lies in $\mathcal{H}_{\mathcal{S}}$ is clearly. We just need to show that $k(\cdot, x_i) -
k(\cdot, \mX_{\mathcal{S}})k(\mX_{\mathcal{S}}, \mX_{\mathcal{S}})^{-1}k(\mX_{\mathcal{S}}, x_i)$ lies in $\mathcal{H}_{\mathcal{S}}^{\perp}$. Notice that
\begin{align}
    &\langle k(\cdot, x_s), k(\cdot, x_i) -
    k(\cdot, \mX_{\mathcal{S}})k(\mX_{\mathcal{S}}, \mX_{\mathcal{S}})^{-1}k(\mX_{\mathcal{S}}, x_i))\rangle_{\mathcal{H}_k}
    \\
    &=
    k(x_s, x_i) -
    k(x_s, \mX_{\mathcal{S}})k(\mX_{\mathcal{S}}, \mX_{\mathcal{S}})^{-1}k(\mX_{\mathcal{S}}, x_i)),\quad\forall (x_s, y_s)\in\mathcal{S}.
\end{align}
If we collect all $\langle k(\cdot, x_s), k(\cdot, x_i) -
k(\cdot, \mX_{\mathcal{S}})k(\mX_{\mathcal{S}}, \mX_{\mathcal{S}})^{-1}k(\mX_{\mathcal{S}}, x_i))\rangle_{\mathcal{H}_k}$ for all $(x_s, y_s) \in \mathcal{S}$, we can obtain
\begin{align}
    k(\mX_{\mathcal{S}}, x_i) -
    k(\mX_{\mathcal{S}}, \mX_{\mathcal{S}})k(\mX_{\mathcal{S}}, \mX_{\mathcal{S}})^{-1}k(\mX_{\mathcal{S}}, x_i))
    =
    k(\mX_{\mathcal{S}}, x_i) - k(\mX_{\mathcal{S}}, x_i) = 0.\label{pf:proj}
\end{align}
This implies that $\langle k(\cdot, x_s), k(\cdot, x_i)-k(\cdot, \mX_{\mathcal{S}})k(\mX_{\mathcal{S}}, \mX_{\mathcal{S}})^{-1}k(\mX_{\mathcal{S}}, x_i))\rangle_{\mathcal{H}_k} =0$ for $x_s \in \mathcal{S}$. $k(\cdot, x_i) -
k(\cdot, \mX_{\mathcal{S}})k(\mX_{\mathcal{S}}, \mX_{\mathcal{S}})^{-1}k(\mX_{\mathcal{S}}, x_i)$ lies in $\mathcal{H}_{\mathcal{S}}^{\perp}$. Eq.~(\ref{pf:proj}) also suggest that $k(x_s, \mX_{\mathcal{S}})k(\mX_{\mathcal{S}}, \mX_{\mathcal{S}})^{-1}k(\mX_{\mathcal{S}}, x_i)$ is equal to $k(x_s, x_i)$ for all $(x_s, y_s)\in\mathcal{S}$. So, $k(\cdot, \mX_{\mathcal{S}})k(\mX_{\mathcal{S}}, \mX_{\mathcal{S}})^{-1}k(\mX_{\mathcal{S}}, x_i)$ is the solution of Eq.~(\ref{Apd:sol:syn}). 
\end{proof}

\subsection{Theorem 1 and Its Proof}\label{sec: Theorem 1 and Its Proof}
\begin{customthm}{1}[Upper bound of conflict loss]
The conflict loss $\mathcal{L}_{\text{conflict}}$ can be bounded as
\begin{align}
    \mathcal{L}_{\text{conflict}} 
    \leq
    \frac{1}{N_A + N_B}\text{Tr}(\mI - k(\mX_{AB}, \mX_{AB}) [k(\mX_{AB}, \mX_{AB}) + (N_A + N_B)\lambda\mI]^{-1})^2 \|\mY_{AB}\|^2_2
\end{align}
where $\text{Tr}$ is the trace operator, $k(\mX_{AB}, \mX_{AB})$ is a $(N_A + N_B)\times (N_A + N_B)$ matrix, and $\mY_{AB}$ is a $(N_A + N_B)\times c$ matrix. 
\end{customthm}
\begin{proof} From Definition~\ref{def:kernel}, we know that the kernel matrix $k(\mX_{AB}, \mX_{AB})$ is positive semidefinite. Hence, there exist some unitary matrix $\mU$ such that $k(\mX_{AB}, \mX_{AB}) = \mU \Sigma \mU^T$ where $\Sigma$ is some diagonal matrix with non-negative components. Then, from Eq.~(\ref{preds:AB}), we can express the upper bound of the conflict loss $\mathcal{L}_{\text{conflict}}$ as
\begin{align}
    \mathcal{L}_{\text{conflict}} 
    &= \frac{1}{N_A + N_B}
    \|\mI\mY_{AB} - \mU \Sigma \mU^T [\mU \Sigma \mU^T + (N_A + N_B)\lambda\mI]^{-1} \mY_{AB}\|_2^2\\
    &= \frac{1}{N_A + N_B}
    \|\mU\mI\mU^T\mY_{AB} - \mU \Sigma \mU^T [\mU (\Sigma + (N_A + N_B)\lambda\mI) \mU^T]^{-1} \mY_{AB}\|_2^2\\
    &= \frac{1}{N_A + N_B}
    \|\mU (\mI - \Sigma [\Sigma + (N_A + N_B)\lambda\mI)]^{-1}) \mU^T\mY_{AB}|^2_2\\
    &= \frac{1}{N_A + N_B}
    \|(\mI - \Sigma [\Sigma + (N_A + N_B)\lambda\mI)]^{-1}) \mU^T\mY_{AB}|^2_2\\
    &\leq \frac{1}{N_A + N_B}
    \|\text{Tr}(\mI - \Sigma [\Sigma + (N_A + N_B)\lambda\mI]^{-1})  \mU^T\mY_{AB}|^2_2\\
    &= \frac{1}{N_A + N_B}
    \text{Tr}(\mI - \Sigma [\Sigma + (N_A + N_B)\lambda\mI]^{-1})^2 \|\mY_{AB}\|^2_2.\label{pf:cf1}
\end{align}
Moreover, we have
\begin{align}
    &\text{Tr}(\mI - k(\mX_{AB}, \mX_{AB}) [k(\mX_{AB}, \mX_{AB}) + (N_A + N_B)\lambda\mI]^{-1}) \nonumber\\
    &= \text{Tr}(\mU (\mI - \Sigma [\Sigma + (N_A + N_B)\lambda\mI]^{-1}) \mU^T) \\
    &= \text{Tr}( (\mI - \Sigma [\Sigma + (N_A + N_B)\lambda\mI]^{-1}) \mU^T\mU) \\
    &= \text{Tr}( (\mI - \Sigma [\Sigma + (N_A + N_B)\lambda\mI]^{-1})).\label{pf:cf2}
\end{align}
Combining Eq.~(\ref{pf:cf1}) and Eq.~(\ref{pf:cf2}) completes the proof.
\end{proof}

\subsection{Theorem 2 and Its Proof}\label{sec: Theorem 2 and Its Proof}
\begin{customthm}{2}[Upper bound of projection loss]
Suppose the kernel matrix of the synthetic dataset $k(\mX_\mathcal{S}, \mX_\mathcal{S})$ is invertible, $f_{\mathcal{S}}$ is the model trained on the synthetic dataset $\mathcal{S}$ with the regularization term $\lambda_{\mathcal{S}}$, where the projection loss $\mathcal{L}_{\text{project}} = \min_{S} \E_{(x, y)\sim \tilde{D}}\ell(f_S, (x, f_{\tilde{D}}(x)))$ can be bounded as
\begin{align}
    \mathcal{L}_{\text{project}} 
    \leq
    \sum_{(x_i, y_i)\in \tilde{D}}
    \min_{\mX_{\mathcal{S}}}
    \sum_{j=1}^c
    \frac{|\alpha_{i, j}|^2}{N_A + N_B}
    \|k(\mX_{AB}, x_i) - k(\mX_{AB}, \mX_\mathcal{S}) 
    k(\mX_\mathcal{S}, \mX_\mathcal{S})^{-1} k(\mX_\mathcal{S}, x_i)\|^2_2.
\end{align}
where $\alpha_{i, j}: = [[k(\mX_{AB}, \mX_{AB}) + (N_A + N_B)\lambda\mI]^{-1} \mY_{AB}]_{i, j}$ ,  which is the weight of $k(\cdot, x_i)$ corresponding to $f^j_{\tilde{D}}$, $\mX_{AB}$ is the $(N_A + N_B) \times d$ matrix corresponding to the features of $\tilde{D}$, $\mX_{\mathcal{S}}$ is the $N_{\mathcal{S}} \times d$ matrix corresponding to the features of $\mathcal{S}$, $\mY_{AB}$ is the $(N_A + N_B) \times c$ matrix corresponding to the labels of $\tilde{D}$, $\mY_{\mathcal{S}}$ is the $N_{\mathcal{S}} \times c$ matrix corresponding to the labels of $\mathcal{S}$.
\end{customthm}

\begin{proof}
From (\ref{sol:AB}), we know that 
\begin{align}
    f_{\tilde{D}}^j(x)
    &=
    [k(x, \mX_{AB}) [k(\mX_{AB}, \mX_{AB}) + (N_A + N_B)\lambda\mI]^{-1} \mY_{AB}]_j\nonumber\\
    &=
    \sum_{(x_i, y_i)\in\tilde{D}} \alpha_{i, j} k(x, x_i).
\end{align}
Then, we can bound the projection loss as
\begin{align}
    &\mathcal{L}_{\text{project}}=
    \min_{\mathcal{S}} \E_{(x, y)\sim \tilde{D}}\ell(f_{\mathcal{S}}, (x, f_{\tilde{D}}(x)))\nonumber\\
    &=
    \min_{\mathcal{S}} \frac{1}{N_A + N_B}\sum_{(x, y)\in\tilde{D}}
    \ell(f_{\mathcal{S}}, (x, f_{\tilde{D}}(x)))\\
    &\leq
    \sum_{(x_i, y_i)\in\tilde{D}}
    \min_{\mathcal{S}} \left\{\frac{1}{N_A + N_B}\sum_{(x, y)\in\tilde{D}}
    \sum_{j=1}^c
    \ell(f_{\mathcal{S}}^j, (x, \alpha_{i, j} k(x, x_i)))\right\}\\
    &=
    \sum_{(x_i, y_i)\in\tilde{D}}
    \min_{\mathcal{S}} \left\{\frac{1}{N_A + N_B}\sum_{(x, y)\in\tilde{D}}
    \sum_{j=1}^c
    \|[k(x, \mX_{\mathcal{S}})[k(\mX_{\mathcal{S}}, \mX_{\mathcal{S}})+N_{\mathcal{S}}\lambda_{\mathcal{S}}\mI]^{-1}\mY_{\mathcal{S}}]_j - \alpha_{i, j} k(x, x_i)\|_2^2\right\}.
\end{align}
For each $(x_i, y_i)\in\tilde{D}$, we have
\begin{align}
&\min_{\mathcal{S}} \left\{\frac{1}{N_A + N_B}\sum_{(x, y)\in\tilde{D}}
\sum_{j=1}^c
\|[k(x, \mX_{\mathcal{S}})[k(\mX_{\mathcal{S}}, \mX_{\mathcal{S}})+N_{\mathcal{S}}\lambda_{\mathcal{S}}\mI]^{-1}\mY_{\mathcal{S}}]_j - \alpha_{i, j} k(x, x_i)\|_2^2\right\}\nonumber\\ 
&\leq
\min_{\mX_{\mathcal{S}}} \left\{\frac{1}{N_A + N_B}\sum_{(x, y)\in\tilde{D}}
\sum_{j=1}^c
\min_{\mY_\mathcal{S}}\|[k(x, \mX_{\mathcal{S}})[k(\mX_{\mathcal{S}}, \mX_{\mathcal{S}})+N_{\mathcal{S}}\lambda_{\mathcal{S}}\mI]^{-1}\mY_{\mathcal{S}}]_j - \alpha_{i, j} k(x, x_i)\|_2^2\right\}\\
&=
\min_{\mX_{\mathcal{S}}} \left\{\frac{1}{N_A + N_B}\sum_{(x, y)\in\tilde{D}}
\sum_{j=1}^c
\min_{f_{i, j}\in\mathcal{H}_{\mathcal{S}}}\|f_{i, j}(x) - \alpha_{i, j} k(x, x_i)\|_2^2\right\}.\label{pf:projloss}
\end{align}
Then, with the help of Lemma~\ref{Apd:lemma:proj}, we bound Eq.~(\ref{pf:projloss}) as follows
\begin{align}
&\min_{\mX_{\mathcal{S}}} \left\{\frac{1}{N_A + N_B}\sum_{(x, y)\in\tilde{D}}
\sum_{j=1}^c
\min_{f_{i, j}\in\mathcal{H}_{\mathcal{S}}}\|f_{i, j}(x) - \alpha_{i, j} k(x, x_i)\|_2^2\right\}\nonumber\\
&\leq
\min_{\mX_{\mathcal{S}}} \left\{\frac{1}{N_A + N_B}\sum_{(x, y)\in\tilde{D}}
\sum_{j=1}^c
\|\alpha_{i,j} [k(x, x_i) - k(x, \mX_{\mathcal{S}})k(\mX_{\mathcal{S}}, \mX_{\mathcal{S}})^{-1}k(\mX_{\mathcal{S}}, x_i)]\|_2^2\right\}\\
&\leq
\min_{\mX_{\mathcal{S}}} \left\{
\sum_{j=1}^c
\frac{|\alpha_{i, j}|^2}{N_A + N_B}
\|k(\mX_{AB}, x_i) - k(\mX_{AB}, \mX_{\mathcal{S}})k(\mX_{\mathcal{S}}, \mX_{\mathcal{S}})^{-1}k(\mX_{\mathcal{S}}, x_i)\|^2_2\right\}\label{pf:projloss2}
\end{align}
We take the summation over $(x_i, y_i)\in\tilde{D}$ for Eq.~(\ref{pf:projloss2}) and then derive the upper bound.
\end{proof}

\subsection{Theorem 3 and Its Proof}\label{sec: Theorem 3 and Its Proof}
\begin{customthm}{3}[Upper bound of generalization gap]
Given a $N$-sample dataset $D$, sampled from the distribution $\mathcal{D}$, then the following generalization gap holds for all $g\in\mathcal{G}$ with probability at least $1-\delta$:
\begin{equation}
    \E_{(x,y)\sim\mathcal{D}}g((x, y)) - \sum_{(x_i, y_i)\in D} \frac{g((x_i, y_i))}{N} 
    \leq 
    2\Hat{\mathfrak{R}}_{D}(\mathcal{G}) 
    + 3 L_{\mathcal{D}} \Gamma_{\mathcal{D}} \sqrt{\frac{\log{\frac{2}{\delta}}}{2N}},
\end{equation}
where $\mX$ is the matrix corresponding to the features of $D$ and $\Hat{\mathfrak{R}}_{D}(\mathcal{G})$ is the empirical Rademacher's complexity.
\end{customthm}

\begin{proof} Here we only sketch the proof, which mainly follows the proof of Theorem 3.3 in \citep{10.5555/2371238}, but is slightly modified under our assumption. First, we denote the maximum of the generalization gap for the dataset $D$ as 
\begin{equation}
    \Phi(D) = \sup_{g\in\mathcal{G}}(\E_{(x, y)\in\mathcal{D}}g((x, y)) - \frac{1}{N}\sum_{(x_i, y_i)\in D} g((x_i, y_i))).
\end{equation}
Consider another dataset $D'$ sampled from the distribution $\mathcal{D}$. $D$ and $D'$ differ by only one sample, which is denoted as $(x_N, y_N)$ and $(x'_N, y'_N)$. Then, according to our assumption, we have
\begin{align}
    \Phi(D) - \Phi(D')
    &\leq
    \sup_{g\in\mathcal{G}} (\frac{1}{N} g((x_N, y_N)) - \frac{1}{N}g((x'_N, y'_N))\\
    &\leq
    \frac{L_D \|(x_N, y_N) - x'_N, y'_N)\|_2}{N}\\
    &\leq
    \frac{L_{\mathcal{D}} \Gamma_{\mathcal{D}}}{N}.
\end{align}
Then, we can apply McDiarmid's inequality on $\Phi(D)$. We can derive
\begin{equation}\label{pf:gen1}
    \Phi(D)\leq\E_{D}\Phi(D) + L_{\mathcal{D}} \Gamma_{\mathcal{D}} \sqrt{\frac{\log{\frac{2}{\delta}}}{2N}},
\end{equation}
which holds with probability at least $1-\frac{\delta}{2}$.
In the proof of Theorem 3.3 in \citep{10.5555/2371238}, we can also prove that $\E_{D}\Phi(D)\leq 2\mathfrak{R}(\mathcal{G})$, where $\mathfrak{R}(\mathcal{G})$ is Rademacher's complexity. Under our assumption, we notice that the empirical Rademacher complexity $\Hat{\mathfrak{R}}_{D}(\mathcal{G})$ also satisfies
\begin{equation}
    \Hat{\mathfrak{R}}_{D}(\mathcal{G}) - \Hat{\mathfrak{R}}_{D'}(\mathcal{G}) 
    \leq
    \frac{L_{\mathcal{D}} \Gamma_{\mathcal{D}}}{N}.
\end{equation}
So, we can apply McDiarmid's inequality again and obtain 
\begin{equation}\label{pf:gen2}
    \mathfrak{R}(\mathcal{G}) \leq \Hat{\mathfrak{R}}_{D}(\mathcal{G}) + L_{\mathcal{D}} \Gamma_{\mathcal{D}} \sqrt{\frac{\log{\frac{2}{\delta}}}{2N}},
\end{equation}
which holds with probability at least $1- \frac{\delta}{2}$. Combine (\ref{pf:gen1}), (\ref{pf:gen2}) and the fact that $\E_{D}\Phi(D)\leq 2\mathfrak{R}(\mathcal{G})$, we have
\begin{equation}
    \E_{(x,y)\sim\mathcal{D}}g((x, y)) - \sum_{(x_i, y_i)\in D} \frac{g((x_i, y_i))}{N} 
    \leq 
    2 \Hat{\mathfrak{R}}_{D}(\mathcal{G})
    + 3 L_{\mathcal{D}} \Gamma_{\mathcal{D}} \sqrt{\frac{\log{\frac{2}{\delta}}}{2N}}.\label{pf:gen3}
\end{equation}
which holds with probability at least $1-\delta$.
\end{proof}

\subsection{Pseudocode for The Simplest Form of KIP-based Backdoor Attack}\label{sec: Pseudocode for The Simplest Form of KIP-based Backdoor Attack}
\begin{algorithm}
\caption{The Simplest Form of KIP-based Backdoor Attack}\label{alg:bdKIP}
\begin{algorithmic}
\Require{benign dataset $D_A$, initial trigger $T_0$, trigger label $y_T$, mask $m$, size of distilled dataset $N_{\mathcal{S}}$, training step $\text{STEP}>0$, batch size $\text{BATCH}>0$, mix ratio $\rho_m > 0$, learning rate $\eta>0$.}
\Ensure synthetic dataset $\mathcal{S}^*$
\State $N \gets 1$
\State $\mathcal{S} \gets$ Randomly sample $N_{\mathcal{S}}$ data from $D_A$ as initial distilled dataset.
\State $D_B\gets\{(x_b, y_b) := ((1-m)\odot x + m\odot T, y_T)| (x_a, y_a)\in D_A\}$
\While{$N \leq \text{STEP}$}

\State $(\mX_A^{\text{batch}}, \mY_A^{\text{batch}}) \gets \text{ Randomly sample BATCH data from } D_A.$

\State $(\mX_B^{\text{batch}}, \mY_B^{\text{batch}}) \gets \text{ Randomly sample BATCH data from } D_B.$

\State $\tilde{D}^{\text{batch}}\gets (\mX_A^{\text{batch}}, \mY_A^{\text{batch}}) \cup (\mX_B^{\text{batch}}, \mY_B^{\text{batch}})$

\State $\mathcal{S} \gets \mathcal{S} - \eta\nabla_{\mathcal{S}} \mathcal{L}(\mathcal{S}, )$
\Comment{$\mathcal{L}$ is defined in Eq.~(\ref{bdKIP:for}).}

\State $N \gets N + 1$
\EndWhile

\State$\mathcal{S}^*\gets\mathcal{S}$
\end{algorithmic}
\end{algorithm}

\subsection{Pseudocode for \textsf{relax-trigger}}\label{sec: Pseudocode for relax-trigger}
\begin{algorithm}
\caption{\textsf{relax-trigger}}\label{alg:triggen}
\begin{algorithmic}
\Require{benign dataset $D_A$, initial trigger $T_0$, trigger label $y_T$, mask $m$, training step $\text{STEP}>0$, batch size $\text{BATCH}>0$, mix ratio $\rho_m > 0$, penalty parameter $\rho>0$, learning rate $\eta>0$.}
\Ensure optimized $T^*$
\State $T \gets T_0$
\State $N \gets 1$
\State $\mathcal{S}_A^* \gets$ Apply KIP to $D_A$ \Comment{We use $\mathcal{S}_A^*$ to denote $\mathcal{S}^*$ from $D_A$}
\While{$N \leq \text{STEP}$}

\State $(\mX_A^{\text{batch}}, \mY_A^{\text{batch}}) \gets \text{ Randomly pick BATCH samples from } D_A$

\State $(\mX^{\text{batch}}, \mY^{\text{batch}}) \gets \text{ Randomly pick } \text{BATCH} \times \rho_m \text{samples from } D_A$ 

\State $(\mX_B^{\text{batch}}, \mY_B^{\text{batch}})\gets\{(x_b, y_b) := ((1-m)\odot x + m\odot T, y_T)| (x, y)\in (\mX^{\text{batch}}, \mY^{\text{batch}})\}$

\State $T \gets T - \eta \nabla_{T}\mathcal{L}(\mathcal{S}_A^*, (\mX_A^{\text{batch}}, \mY_A^{\text{batch}}), (\mX_B^{\text{batch}}, \mY_B^{\text{batch}}), \rho)$ \Comment{$\mathcal{L}$ is defined in Eq.~(\ref{eq:triggen}).}

\State $N \gets N + 1$
\EndWhile

\State $T^* \gets T$
\end{algorithmic}
\end{algorithm}

\subsection{Extra Experiments.}\label{sec: Extra Experiments}
In Tables~\ref{tab:def:cifar10-500}$\sim$ \ref{tab:def:relax-trig}, we provide extra experimental results.

\begin{table}[ht!]
    \centering
    \resizebox{\textwidth}{!}{
    \begin{tabular}{|c||c|c|c|c|c|c|c|c|c|c|c|c|c|c|}
    \hline
    $\text{Trig.}\backslash\text{Def.}$ & \multicolumn{2}{|c|}{None} & \multicolumn{2}{|c|}{SCAn} & \multicolumn{2}{|c|}{AC} & \multicolumn{2}{|c|}{SS} & \multicolumn{2}{|c|}{Strip}\\
    \hline
    \multicolumn{1}{|c||}{} & \multicolumn{1}{|c}{CTA (\%)} & \multicolumn{1}{c|}{ASR (\%)} & 
    \multicolumn{1}{|c}{CTA (\%)} & \multicolumn{1}{c|}{ASR (\%)} & \multicolumn{1}{|c}{CTA (\%)} & \multicolumn{1}{c|}{ASR (\%)} & \multicolumn{1}{|c}{CTA (\%)} & \multicolumn{1}{c|}{ASR (\%)} & \multicolumn{1}{|c}{CTA (\%)} & \multicolumn{1}{c|}{ASR (\%)} \\
    \hline
    $2\times 2$ & 29.47 (0.44)& 25.67 (4.35) & 28.70 (2.23) & 35.29 (4.40) & 29.93 (1.66)& 26.17 (9.14) & 28.13 (1.45) & 19.24 (4.39) & 27.52 (3.30) & 28.73 (13.83)\\ 
    \hline
    $4\times 4$ & 29.72 (1.62) & 31.08 (8.07) & 32.43 (1.87) & 26.70 (2.30) & 30.57 (0.87) & 30.20 (3.17) & 30.48 (0.91) & 11.07 (1.53)	& 25.95 (4.07) & 25.37 (11.61)\\ 
    \hline
    $8\times 8$ & 32.00 (1.03) & 51.65 (12.41) & 30.78 (2.08) & 37.45 (9.23) & 29.57 (0.74) & 35.99 (2.63) & 28.46 (2.56) & 12.77 (3.79) & 28.39 (3.18) & 15.94 (2.67)\\ 
    \hline
    $16\times 16$ & 34.61 (1.01) & 85.65 (17.12) & 33.88 (1.65) & 44.24 (3.85) & 31.96 (0.53) & 59.56 (21.29) & 30.70 (0.09) & 44.77 (25.11) &	27.96 (7.42) & 43.00 (25.69)\\ 
    \hline
    $32\times 32$ & 33.78 (0.53) & 100.00 (0.00) &	34.54 (0.93) & 100.00 (0.00) & 32.25 (2.29) & 33.33 (47.14) & 29.04 (0.91) & 33.33 (47.14)	& 26.93 (8.95) & 0.00 (0.00)\\ 
    \hline\hline
    $\text{Trig.}\backslash\text{Def.}$ & \multicolumn{2}{|c|}{} & \multicolumn{2}{|c|}{ABL} & \multicolumn{2}{|c|}{NAD} & \multicolumn{2}{|c|}{STRIP} & \multicolumn{2}{|c|}{FP}\\
    \hline
    \multicolumn{1}{|c||}{} & \multicolumn{1}{|c}{} & \multicolumn{1}{c|}{} & 
    \multicolumn{1}{|c}{CTA (\%)} & \multicolumn{1}{c|}{ASR (\%)} & \multicolumn{1}{|c}{CTA (\%)} & \multicolumn{1}{c|}{ASR (\%)} & \multicolumn{1}{|c}{CTA (\%)} & \multicolumn{1}{c|}{ASR (\%)} & \multicolumn{1}{|c}{CTA (\%)} & \multicolumn{1}{c|}{ASR (\%)} \\
    \hline
    $2\times 2$ 
    & & & 25.31 (4.40) & 13.67 (6.65) & 36.27 (0.54) & 7.32 (1.77) & 25.05 (2.73) & 27.25 (8.88) & 15.28 (2.96) & 35.74 (41.58)\\		
    \hline
    $4\times 4$ 
    & & & 24.25 (2.70) & 5.81 (5.26) & 36.29 (2.07) & 6.57 (0.81) & 26.85 (1.39) & 27.92 (7.10) & 14.67 (3.67) & 69.41 (31.97)\\
    \hline
    $8\times 8$ 
    & & & 22.97 (5.54) & 19.06 (8.02) & 36.96 (1.73) & 14.53 (5.15) & 28.84 (0.89) & 40.99 (8.03) & 19.41 (2.06) & 75.62 (17.01)\\
    \hline
    $16\times 16$ 
    & & & 28.43 (2.31) & 64.73 (34.70) & 37.01 (1.16) & 29.42 (1.39) & 31.09 (0.80) & 22.80 (22.02) & 21.25 (2.94) & 19.08 (21.95)\\
    \hline
    $32\times 32$ 
    & & & 22.12 (2.74) & 66.67 (47.14) & 32.62 (7.23) & 66.67 (47.14) & 22.99 (10.01) & 0.00 (0.00) & 17.67 (5.61) & 33.33 (47.14)\\
    \hline
    \end{tabular}}
    \caption{Defenses for \textsf{simple-trigger} on CIFAR-10 with size $500$.}
    \label{tab:def:cifar10-500}
        
\end{table}

\begin{table}[t]
    \centering
    \resizebox{\textwidth}{!}{
    \begin{tabular}{|c||c|c|c|c|c|c|c|c|c|c|c|c|c|c|c|c|c|c|}
    \hline
     $\text{Trig.}\backslash\text{Def.}$ & \multicolumn{2}{|c|}{None} & \multicolumn{2}{|c|}{SCAn} & \multicolumn{2}{|c|}{AC} & \multicolumn{2}{|c|}{SS} & \multicolumn{2}{|c|}{Strip} \\
    \hline
    \multicolumn{1}{|c||}{} & \multicolumn{1}{|c}{CTA (\%)} & \multicolumn{1}{c|}{ASR (\%)} & 
    \multicolumn{1}{|c}{CTA (\%)} & \multicolumn{1}{c|}{ASR (\%)} & \multicolumn{1}{|c}{CTA (\%)} & \multicolumn{1}{c|}{ASR (\%)} & \multicolumn{1}{|c}{CTA (\%)} & \multicolumn{1}{c|}{ASR (\%)} & \multicolumn{1}{|c}{CTA (\%)} & \multicolumn{1}{c|}{ASR (\%)}\\
    \hline
    $2\times 2$ & 37.78 (2.09) & 10.71 (2.33) & 40.17 (8.89) & 5.18 (2.40) & 27.61 (1.09) & 9.43 (4.63) & 14.55 (1.71) & 6.47 (3.30) & 28.52 (2.29) & 5.22 (4.19) \\
    \hline
    $4\times 4$ & 39.07 (2.27) & 18.67 (8.87) & 33.47 (6.75) & 3.59 (1.11) & 23.22 (7.42) & 9.71 (5.61) & 15.22 (2.03) & 2.28 (2.36)  & 26.56 (4.15) & 7.67 (2.29)\\
    \hline
    $8\times 8$ & 38.89 (1.68) & 47.53 (9.30) & 21.48 (12,74) & 2.85 (0.37) & 28.60 (1.36) & 8.25 (4.15) & 12.86 (1.78) & 8.60 (6.45) & 34.96 (5.02) & 8.69 (1.85)\\
    \hline
    $16\times 16$ & 37.92 (2.84) & 84.24 (3.60) & 32.99 (4.62) & 11.04 (8.85) & 26.91 (2.77) & 56.51 (39.03) & 14.01 (1.84) & 4.76 (4.32) & 29.22 (1.59) & 37.06 (41.29) \\
    \hline
    $32\times 32$ & 41.97 (0.97) & 66.67 (47.14) & 33.47 (9.14) & 33.33 (47.14) & 27.20 (3.83) & 33.33 (47.14) & 13.99 (0.79) & 33.33 (47.14) & 35.07 (1.95) & 33.33 (47.14)\\
    \hline\hline
    $\text{Trig.}\backslash\text{Def.}$ & \multicolumn{2}{|c|}{} 
    & \multicolumn{2}{|c|}{ABL} & \multicolumn{2}{|c|}{NAD} & \multicolumn{2}{|c|}{STRIP} & \multicolumn{2}{|c|}{FP}\\
    \hline
    \multicolumn{1}{|c||}{} & \multicolumn{1}{|c}{} & \multicolumn{1}{c|}{} & 
    \multicolumn{1}{|c}{CTA (\%)} & \multicolumn{1}{c|}{ASR (\%)} & \multicolumn{1}{|c}{CTA (\%)} & \multicolumn{1}{c|}{ASR (\%)} & \multicolumn{1}{|c}{CTA (\%)} & \multicolumn{1}{c|}{ASR (\%)} & \multicolumn{1}{|c}{CTA (\%)} & \multicolumn{1}{c|}{ASR (\%)}\\
    \hline
    $2\times 2$ 
    & & & 32.31 (4.67) & 5.85 (1.90) & 94.05 (0.51) & 0.40 (0.04) & 22.36 (14.79) &	5.48 (4.00) & 14.57 (10.65) & 8.13 (5.75)\\			
    \hline
    $4\times 4$ 
    & & & 36.25 (4.05) & 6.68 (3.75) & 93.75 (0.41) & 0.29 (0.09) & 35.23 (2.17) & 16.99 (8.05) & 23.73 (4.09) & 11.45 (11.59)\\	
    \hline
    $8\times 8$ 
    & & & 25.49 (15.17) & 11.37 (3.68) & 93.72 (0.11) & 0.45 (0.19) & 34.82 (1.63) & 42.69 (7.99) & 24.67 (1.23) & 33.71 (29.71)\\
    \hline
    $16\times 16$ 
    & & & 37.15 (1.59) & 63.80 (31.74) & 93.78 (0.07) & 0.09 (0.08) &	34.04 (2.48) & 50.69 (10.77) & 25.35 (2.59) & 40.16 (30.45)\\	
    \hline
    $32\times 32$ 
    & & & 34.56 (5.89) & 33.33 (47.14) & 94.05 (0.17) & 0.00 (0.00) &	37.87 (0.80) & 0.00 (0.00) & 25.43 (2.06) & 66.67 (47.14)\\
    \hline
    \end{tabular}}
    \caption{Defenses for \textsf{simple-trigger} on GTSRB with distilled dataset size = $430$.}
    \label{tab:def:gtsrb-430}
\end{table}

\begin{table}[H]
    \centering
    \resizebox{\textwidth}{!}{
    \begin{tabular}{|c||c|c|c|c|c|c|c|c|c|c|c|c|c|c|c|c|c|c|}
    \hline
    $\text{Trig.}\backslash\text{Def.}$ & \multicolumn{2}{|c|}{None} & \multicolumn{2}{|c|}{SCAn} & \multicolumn{2}{|c|}{AC} & \multicolumn{2}{|c|}{SS} & \multicolumn{2}{|c|}{Strip} \\
    \hline
    \multicolumn{1}{|c||}{} & \multicolumn{1}{|c}{CTA (\%)} & \multicolumn{1}{c|}{ASR (\%)} & 
    \multicolumn{1}{|c}{CTA (\%)} & \multicolumn{1}{c|}{ASR (\%)} & \multicolumn{1}{|c}{CTA (\%)} & \multicolumn{1}{c|}{ASR (\%)} & \multicolumn{1}{|c}{CTA (\%)} & \multicolumn{1}{c|}{ASR (\%)} & \multicolumn{1}{|c}{CTA (\%)} & \multicolumn{1}{c|}{ASR (\%)}\\
    \hline
    $2\times 2$ & 72.23 (2.76) & 1.03 (0.42) & 72.82 (12.40) & 2.37 (1.69) & 63.73 (7.87) & 2.06 (0.56) & 44.50 (2.26) & 3.08 (0.81) & 76.63 (3.00) & 1.00 (0.47) \\
    \hline
    $4\times 4$ & 73.29 (1.22) & 1.08 (0.37) & 81.19 (2.30) & 1.18 (0.09) & 71.79 (1.78) & 2.37 (0.40) & 45.16 (5.15) & 3.53 (1.12) & 73.28 (11.33)  & 0.94 (0.46)\\
    \hline
    $8\times 8$ & 73.29 (0.26) & 8.08 (4.20) & 79.28 (2.69) & 3.84 (1.72) & 62.79 (9.81) & 6.30 (3.27) & 40.46 (4.65) & 17.40 (7.28) & 74.84 (1.37) & 2.78 (1.41) \\
    \hline
    $16\times 16$ & 73.12 (0.69) & 70.10 (13.96) & 81.39 (5.97) & 61.28 (19.18) & 68.37 (2.70) &	46.99 (24.09) & 39.58 (0.15) & 22.70 (11.70) & 73.52 (6.41) & 28.07 (18.36) \\
    \hline
    $32\times 32$ & 74.13 (1.39) & 100.00 (0.00) & 76.85 (5.79) & 100.00 (0.00) & 45.93 (21.02)	& 33.33 (47.14) & 44.87 (6.65) & 33.33 (47.14) & 83.42 (0.65) & 0.00 (0.00) \\
    \hline\hline
    $\text{Trig.}\backslash\text{Def.}$ & \multicolumn{2}{|c|}{} 
    & \multicolumn{2}{|c|}{ABL} & \multicolumn{2}{|c|}{NAD} & \multicolumn{2}{|c|}{STRIP} & \multicolumn{2}{|c|}{FP}\\
    \hline
    \multicolumn{1}{|c||}{} & 
    \multicolumn{1}{|c}{} & \multicolumn{1}{c|}{} & 
    \multicolumn{1}{|c}{CTA (\%)} & \multicolumn{1}{c|}{ASR (\%)} & \multicolumn{1}{|c}{CTA (\%)} & \multicolumn{1}{c|}{ASR (\%)} & \multicolumn{1}{|c}{CTA (\%)} & \multicolumn{1}{c|}{ASR (\%)} & \multicolumn{1}{|c}{CTA (\%)} & \multicolumn{1}{c|}{ASR (\%)}\\
    \hline
    $2\times 2$ 
     & & & 77.04 (4.01) & 1.86 (2.06) & 94.68 (0.60) & 0.32 (0.05) & 65.05 (2.53) & 0.95 (0.39) & 51.95 (2.11) & 0.29 (0.17) \\
    \hline
    $4\times 4$
    & & & 79.11 (1.23) & 1.04 (0.93) & 94.73 (0.26) & 0.27 (0.04) &	65.94 (1.11) & 1.02 (0.34) & 52.61 (0.49) & 0.04 (0.06) \\
    \hline
    $8\times 8$ 
    & & & 75.89 (1.80) & 4.14 (2.19) & 94.76 (0.33) & 0.24 (0.07) &	65.90 (0.23) & 7.65 (4.03) & 51.83 & 0.23 \\
    \hline
    $16\times 16$ 
    & & & 79.29 (3.46)  & 73.46 (3.25)  & 94.67 (0.29)  & 0.07 (0.04)  & 65.73 (0.65)  & 59.97 (5.28)  & 53.38 (3.29)  & 28.53 (39.26)  \\	
    \hline
    $32\times 32$ 
    & & & 79.98 (2.47)  & 100.00 (0.00)  & 94.74 (0.25)  & 0.00 (0.00)  & 66.74 (1.34)  & 0.00 (0.00)  & 53.79 (2.14)  &	100.00 (0.00)\\
    \hline
    \end{tabular}}
    \caption{Defenses for \textsf{simple-trigger} on GTSRB with distilled dataset size $2150$.}
    \label{tab:def:gtsrb-2150}
       
\end{table}

\begin{table}[H]
    \centering
    \resizebox{\textwidth}{!}{
    \begin{tabular}{|c||c|c|c|c|c|c|c|c|c|c|c|c|c|c|c|c|c|c|}
    \hline
    $\text{Data. (Size)}\backslash\text{Def.}$ & \multicolumn{2}{|c|}{None} & \multicolumn{2}{|c|}{SCAn} & \multicolumn{2}{|c|}{AC} & \multicolumn{2}{|c|}{SS} & \multicolumn{2}{|c|}{Strip} \\
    \hline
    \multicolumn{1}{|c||}{} & \multicolumn{1}{|c}{CTA (\%)} & \multicolumn{1}{c|}{ASR (\%)} & 
    \multicolumn{1}{|c}{CTA (\%)} & \multicolumn{1}{c|}{ASR (\%)} & \multicolumn{1}{|c}{CTA (\%)} & \multicolumn{1}{c|}{ASR (\%)} & \multicolumn{1}{|c}{CTA (\%)} & \multicolumn{1}{c|}{ASR (\%)} & \multicolumn{1}{|c}{CTA (\%)} & \multicolumn{1}{c|}{ASR (\%)}\\
    \hline
    CIFAR-10 (100) & 26.28 (1.56) & 42.10 (4.16) & 27.23 (1.37) & 55.63 (4.78 & 26.18 (1.70) & 44.56 (13.65)  & 21.13 (0.67)  & 5.98 (0.96)  & 25.05 (0.79)  & 50.27 (15.74) \\
    \hline
    CIFAR-10 (500)  & 33.98 (0.63)  & 90.87 (4.01)  & 35.40 (0.77)  & 82.63 (3.21)  & 33.23 (2.63)  & 58.22 (38.26)  & 32.52 (0.47)  & 29.00 (10.45) & 34.33 (0.23) & 83.42 (6.34) \\
    \hline
    GTSRB (430) & 37.49 (1.98) & 69.14 (2.84) &	36.86 (1.28) & 53.68 (4.41) & 23.91 (2.58) & 28.27 (11.06) & 13.39 (1.19) & 25.09 (12.68) & 30.88 (2.18) & 50.50 (13.23) \\
    \hline
    GTSRB (2150) & 75.40 (0.39) & 65.28 (2.15) &	82.47 (1.81) & 70.51 (3.14) & 65.84 (8.99) & 61.81 (1.22) & 39.95 (3.48) & 26.01 (11.47) & 72.43 (5.26) & 63.77 (1.18) \\
    \hline\hline
    $\text{Data. (Size)}\backslash\text{Def.}$ & \multicolumn{2}{|c|}{} 
    & \multicolumn{2}{|c|}{ABL} & \multicolumn{2}{|c|}{NAD} & \multicolumn{2}{|c|}{STRIP} & \multicolumn{2}{|c|}{FP}\\
    \hline
    \multicolumn{1}{|c||}{} & 
    \multicolumn{1}{|c}{} & \multicolumn{1}{c|}{} & 
    \multicolumn{1}{|c}{CTA (\%)} & \multicolumn{1}{c|}{ASR (\%)} & \multicolumn{1}{|c}{CTA (\%)} & \multicolumn{1}{c|}{ASR (\%)} & \multicolumn{1}{|c}{CTA (\%)} & \multicolumn{1}{c|}{ASR (\%)} & \multicolumn{1}{|c}{CTA (\%)} & \multicolumn{1}{c|}{ASR (\%)}\\
    \hline
    CIFAR-10 (100) 
     & & & 14.03 (0.92) & 73.30 (17.71) & 31.60 (2.10) & 21.67 (18.72) & 23.83 (1.30) & 35.96 (5.88) & 15.87 (1.08) & 33.50 (38.47)\\
    \hline
    CIFAR-10 (500) 
    & & & 29.10 (1.51) & 13.41 (5.42) & 37.75 (1.19) & 44.20 (10.85) & 30.61 (0.49) & 40.13 (18.51) & 20.83 (1.43) & 32.28 (44.66) \\
    \hline
    GTSRB (430) 
    & & & 32.26 (2.68) & 45.02 (6.54) & 93.32 (0.34) & 67.18 (2.36) & 33.88 (1.82) & 61.89 (2.63) & 22.79 (2.98) & 53.54 (13.43)\\
    \hline
    GTSRB (2150) 
    & & & 80.90 (1.39) & 65.85 (0.63) & 94.34 (0.13) & 33.68 (1.43) & 67.91 (0.31) & 45.40 (2.13) & 55.81 (1.39) & 68.73 (0.57) \\
    \hline
    \end{tabular}}
    \caption{Defenses for \textsf{relax-trigger} on CIFAR-10 and GTSRB.}
    \label{tab:def:relax-trig}
        \vspace{-4mm}
\end{table}

\section{Ablation Studies}
{
\subsection{KIP-based backdoor attack on ImageNet}
We perform our KIP-based backdoor attack on ImageNet. In our experiment, we randomly choose ten sub-classes to perform our experiment. We also resize each image in the ImageNet into 128x128. The experimental results show that our KIP-based backdoor attack is effective (see Table~\ref{tab:abl:exp1}).
\begin{table}[!ht]
    \centering
    \begin{tabular}{|l|l|l|l|l|l|}
    \hline
        Trigger-type & Dataset & Model & IPC (Image Per Class) & CTA (\%) & ASR (\%) \\ \hline
        \textsf{simple-trigger} & ImageNet & NTK & 10 & 15.00 & 100.00 \\ \hline
        \textsf{simple-trigger} & ImageNet & NTK & 50 & 16.60 & 100.00 \\ \hline
        \textsf{relax-trigger} & ImageNet & NTK & 10 & 16.40 & 100.00 \\ \hline
        \textsf{relax-trigger} & ImageNet & NTK & 50 & 17.00 & 100.00 \\ \hline
    \end{tabular}
    \caption{Efficacy of KIP-based backdoor attack on ImageNet.}
    \label{tab:abl:exp1}
\end{table}
}

{
\subsection{Impact of IPC on KIP-based Backdoor Attack}
We examine the efficacy of KIP-based backdoor attack influenced by IPC (Image Per Class). We examine the efficacy of \textsf{simple-trigger} and \textsf{relax-trigger} under different sizes of synthetic dataset (IPC 10 $\sim$ IPC 50). The experimental results show that both CTA and ASR are gradually rising as the IPC increases (see Table~\ref{tab:abl:exp2_1}). The corresponding experiments for DOORPING is presented in Table~\ref{tab:abl:exp2_2}.

\begin{table}[!ht]
    \centering
    \begin{tabular}{|l|l|l|l|l|}
    \hline
        Dataset & Trigger-type & IPC (Image Per Class) & CTA (\%) & ASR (\%) \\ \hline
        CIFAR-10 & simple-trigger & 10 & 41.70 (0.25) & 100 (0.00) \\ \hline
        CIFAR-10 & simple-trigger & 20 & 42.58 (0.23) & 100 (0.00) \\ \hline
        CIFAR-10 & simple-trigger & 30 & 43.29 (0.35) & 100 (0.00) \\ \hline
        CIFAR-10 & simple-trigger & 40 & 43.55 (0.42) & 100 (0.00) \\ \hline
        CIFAR-10 & simple-trigger & 50 & 43.66 (0.40) & 100 (0.00) \\ \hline
        ~ & ~ & ~ & ~ & ~ \\ \hline
        CIFAR-10 & relax-trigger & 10 & 41.66 (0.01) & 100 (0.00) \\ \hline
        CIFAR-10 & relax-trigger & 20 & 42.46 (0.01) & 100 (0.00) \\ \hline
        CIFAR-10 & relax-trigger & 30 & 42.99 (0.08) & 100 (0.00) \\ \hline
        CIFAR-10 & relax-trigger & 40 & 43.10 (0.09) & 100 (0.00) \\ \hline
        CIFAR-10 & relax-trigger & 50 & 43.64 (0.40) & 100 (0.00) \\ \hline
        ~ & ~ & ~ & ~ & ~ \\ \hline
        GTSRB & simple-trigger & 10 & 67.56 (0.60) & 100 (0.00) \\ \hline
        GTSRB & simple-trigger & 20 & 69.44 (0.35) & 100 (0.00) \\ \hline
        GTSRB & simple-trigger & 30 & 70.24 (0.38) & 100 (0.00) \\ \hline
        GTSRB & simple-trigger & 40 & 70.84 (0.32) & 100 (0.00) \\ \hline
        GTSRB & simple-trigger & 50 & 71.27 (0.24) & 100 (0.00) \\ \hline
        ~ & ~ & ~ & ~ & ~ \\ \hline
        GTSRB & relax-trigger & 10 & 68.73 (0.67) & 95.26 (0.54) \\ \hline
        GTSRB & relax-trigger & 20 & 70.38 (0.03) & 94.85 (0.13) \\ \hline
        GTSRB & relax-trigger & 30 & 71.26 (0.02) & 95.73 (0.32) \\ \hline
        GTSRB & relax-trigger & 40 & 71.81 (0.01) & 95.84 (0.18) \\ \hline
        GTSRB & relax-trigger & 50 & 71.54 (0.33) & 95.08 (0.33) \\ \hline
    \end{tabular}
    \caption{Efficacy of KIP-based backdoor attack influenced by the size of the synthetic dataset.}
    \label{tab:abl:exp2_1}
\end{table}

\begin{table}[!ht]
    \centering
    \begin{tabular}{|l|l|l|l|l|}
    \hline
    Dataset & Trigger-type & IPC (Image Per Class) & CTA (\%) & ASR   (\%) \\ \hline
    CIFAR-10 & DOORPING & 10 & 36.35 (0.42) & 80.00 (40.00) \\ \hline
    CIFAR-10 & DOORPING & 20 & 37.65 (0.42) & 70.00 (45.83) \\ \hline
    CIFAR-10 & DOORPING & 30 & 38.48 (0.36) & 90.00 (30.00) \\ \hline
    CIFAR-10 & DOORPING & 40 & 37.78 (0.61) & 70.00 (45.83) \\ \hline
    GTSRB & DOORPING & 10 & 68.03 (0.92) & 90.00 (30.00) \\ \hline
    GTSRB & DOORPING & 20 & 81.45 (0.46) & 80.00 (40.00) \\ \hline
    GTSRB & DOORPING & 30 & 81.62 (0.71) & 100.00 (0.00) \\ \hline
    \end{tabular}
    \caption{Efficacy of DOORPING influenced by the size of the synthetic dataset.}
    \label{tab:abl:exp2_2}
\end{table}
}

{
\subsection{Cross Model Ability of KIP-based backdoor attack}

The experiment for cross model ability is presented in Table~\ref{tab:abl:exp3}. We train the distilled dataset poinsoned by \textsf{simple-trigger} and \textsf{relax-trigger} on 3-layers MLP and 3-layers ConvNet. The experimental results show that both CTA and ASR go up as we increase the IPC (Image Per Class), which suggests that the cross model issue may be relieved as the IPC is large enough.
\begin{table}[!ht]
    \centering
    \begin{tabular}{|l|l|l|l|l|l|}
    \hline
        Dataset & Trigger-type & IPC (Image Per Class) & Cross\_model & CTA (\%) & ASR(\%) \\ \hline
        CIFAR-10 & \textsf{simple-trigger} & 10 & MLP & 11.58 (2.10) & 40.00 (48.98)  \\ \hline
        CIFAR-10 & \textsf{simple-trigger} & 10 & CNN & 47.37 (7.44) & 40.00 (48.98)  \\ \hline
        CIFAR-10 & \textsf{simple-trigger} & 10 & NTK (baseline) & 41.70 (0.25) & 100.00 (0.00)  \\ \hline
        ~ & ~ & ~ & ~ & ~ & ~  \\ \hline
        CIFAR-10 & \textsf{simple-trigger} & 50 & MLP & 48.08 (4.72) & 40.00 (48.98)  \\ \hline
        CIFAR-10 & \textsf{simple-trigger} & 50 & CNN & 95.96 (1.10) & 100.00 (0.00)  \\ \hline
        CIFAR-10 & \textsf{simple-trigger} & 50 & NTK (baseline) & 43.66 (0.40) & 100.00 (0.00)  \\ \hline
        ~ & ~ & ~ & ~ & ~ & ~  \\ \hline
        CIFAR-10 & \textsf{relax-trigger} & 10 & MLP & 10.52 (7.44) & 19.40 (38.80)  \\ \hline
        CIFAR-10 & \textsf{relax-trigger} & 10 & CNN & 64.21 (6.98) & 81.80 (11.44)  \\ \hline
        CIFAR-10 & \textsf{relax-trigger} & 10 & NTK (baseline) & 41.66 (0.74) & 100.00 (0.00)  \\ \hline
        ~ & ~ & ~ & ~ & ~ & ~  \\ \hline
        CIFAR-10 & \textsf{relax-trigger} & 50 & MLP & 44.24 (4.49) & 78.28 (24.57)  \\ \hline
        CIFAR-10 & \textsf{relax-trigger} & 50 & CNN & 93.13 (2.24) & 82.80 (6.53)  \\ \hline
        CIFAR-10 & \textsf{relax-trigger} & 50 & NTK (baseline) & 43.64 (0.40) & 100.00 (0.00)  \\ \hline
    \end{tabular}
    \caption{Experiment of cross model ability of KIP-based backdoor attack.}
    \label{tab:abl:exp3}
\end{table}
}

{
\subsection{Transferability of KIP-based backdoor attack}
Our KIP-based backdoor attack can evade other data distillation techniques. In particular, we perform experiments to examine the transferability of our theory-induced triggers. We first use our \textsf{simple-trigger} and \textsf{relax-trigger} to poison the dataset. Then, we distill dataest with a different distillation method, FRePo~\citep{zhou2022dataset} and DM~\citep{zhao2023dataset}. The experimental results shows that our triggers can successfully transfer to the FrePo and DM (see Table~\ref{tab:abl:exp4_1} and Table~
\ref{tab:abl:exp4_2}).

\begin{table}[!ht]
    \centering
    \begin{tabular}{|l|l|l|l|l|l|l|}
    \hline
        Trigger-type & Dataset & IPC (Image Per Class) & Distillation & Model & CTA (\%) & ASR (\%) \\ \hline
        CIFAR-10 & \textsf{simple-trigger} & 10 & FRePO & ConvNet & 60.32 & 83.10 \\ \hline
        CIFAR-10 & \textsf{relax-trigger} & 50 & FRePO & ConvNet & 68.34 & 81.61 \\ \hline
    \end{tabular}
    \caption{Experiment of transferability for FRePO.}
    \label{tab:abl:exp4_1}
\end{table}
\begin{table}[!ht]
    \centering
    \begin{tabular}{|l|l|l|l|l|l|l|}
    \hline
        Trigger-type & Dataset & IPC (Image Per Class) & Distillation & Model & CTA (\%) & ASR (\%) \\ \hline
        Cifar10 & simple-trigger & 10 & DM & MLP & 36.41 & 77.03 \\ \hline
        Cifar10 & simple-trigger & 50 & DM & MLP & 36.88 & 76.79 \\ \hline
        Cifar10 & relax-trigger & 10 & DM & MLP & 36.31 & 76.04 \\ \hline
        Cifar10 & relax-trigger & 50 & DM & MLP & 36.81 & 76.21 \\ \hline
    \end{tabular}
    \caption{Experiment of transferability for DM.}
    \label{tab:abl:exp4_2}
\end{table}
}

{
\subsection{KIP-based Backdoor Attack on NAS and CL}
We train our distilled dataset poinsoned by \textsf{simple-trigger} and \textsf{relax-trigger} in different scenarios, neural architecture search (NAS) and continual learning (CL). The experimental results are shown in Table~\ref{tab:abl:exp5_1} and Table~\ref{tab:abl:exp5_2}.
\begin{table}[!ht]
    \centering
    \begin{tabular}{|l|l|l|l|l|l|}
    \hline
        Trigger-type & Dataset & IPC (Image Per Class) & Scenario & CTA (\%) & ASR (\%) \\ \hline
        simple-trigger & CIFAR-10 & 50 & NAS & 37.49(3.44) & 100.00(0.00) \\ \hline
        relax-trigger & CIFAR-10 & 50 & NAS & 36.43(3.62) & 86.23(3.22) \\ \hline
    \end{tabular}
    \caption{Experiment for NAS. The experiment result shows that our triggers remain effective for NAS.}
    \label{tab:abl:exp5_1}
\end{table}

\begin{table}[!ht]
    \centering
    \begin{tabular}{|l|l|l|l|l|l|}
    \hline
        Trigger-type & Dataset & IPC (Image Per Class) & Scenario & CTA (\%) & ASR (\%) \\ \hline
        simple-trigger & CIFAR-10 & 50 & CL & 13.93(1.93) & 100.00(0.00) \\ \hline
        simple-trigger & CIFAR-10 & 50 & baseline & 13.60(1.66) & 100.00(0.00) \\ \hline
        relax-trigger & CIFAR-10 & 50 & CL & 20.13(2.94) & 60.94(21.68) \\ \hline
        relax-trigger & CIFAR-10 & 50 & baseline & 14.00(3.54) & 43.11(7.83) \\ \hline
    \end{tabular}
    \caption{Experiment for CL. The experiment result shows that both CTA and ASR are slightly higher than baseline.}
    \label{tab:abl:exp5_2}
\end{table}

Note that the details about our implementation of NAS and CL are described below. 

\begin{itemize}
    \item[\textbf{NAS}]: The process defines a search space (random search) that includes a range of possible model parameters such as the number of convolutional layers, the number of dense layers, and the size of the convolutional layers. The program randomly selects parameters from this space to generate multiple candidate model architectures. A CNN model is then built, comprising convolutional layers (Conv2D), batch normalization (BatchNormalization), activation functions (such as ReLU), pooling layers (MaxPooling2D), flattening layers (Flatten), fully connected layers (Dense), and optionally Dropout layers. Each model is compiled and trained using the Adam optimizer and categorical cross-entropy loss function, but in this case, the same dataset is used for evaluation (although typically, an independent test set should be used). The accuracy and loss functions of different models are compared, and ultimately the best model is selected and saved
    \item[\textbf{CL}]: The dataset is divided into different category-specific subsets (as in CIFAR-10, which is divided into 10 categories), each containing images and their corresponding labels. This allows the model to gradually train on each subset. A CNN model is built, including multiple convolutional layers (Conv2D), batch normalization layers (BatchNormalization), ReLU activation functions, max pooling layers (MaxPooling2D), and fully connected layers (Dense). The final layer uses a softmax activation function, a typical configuration for label classification tasks. The model is compiled using an RMSprop optimizer and categorical cross-entropy loss function. Further training optimization can be applied, such as using Elastic Weight Consolidation (EWC) to minimize the impact on the originally trained model when learning new subsets.
\end{itemize}
}

{
\subsection{Performance of the Triggers without Distillation}
We perform the experiments on CIFAR-10 and GTSRB. We first utilize the \textsf{simple-trigger} and \textsf{relax-trigger} to poison the dataset. Then, we use 3-layers ConvNet to train a model and evaluate corresponding CTA and ASR.
The experimental results demonstrate that our triggers \textsf{simple-trigger} and \textsf{relax-trigger} both remain effective (see Table~\ref{tab:abl:exp6}).
\begin{table}[!ht]
    \centering
    \begin{tabular}{|l|l|l|l|l|}
    \hline
        Dataset & Trigger-type & Transparency (m) & CTA (\%) & ASR (\%) \\ \hline
        CIFAR-10 & simple-trigger & 1 & 70.02 (0.40) & 100.00 (0.00) \\ \hline
        CIFAR-10 & relax-trigger & 0.3 & 70.02 (0.65) & 99.80 (0.04) \\ \hline
        CIFAR-10 & simple-trigger & 0.3 & 67.84 (0.36) & 95.50 (1.23) \\ \hline
        ~ & ~ & ~ & ~ & ~ \\ \hline
        GTSRB & simple-trigger & 1 & 72.47 (3.36) & 100.00 (0.00) \\ \hline
        GTSRB & relax-trigger & 0.3 & 75.50 (2.09) & 99.82 (0.09) \\ \hline
        GTSRB & simple-trigger & 0.3 & 70.21 (3.03) & 99.36 (0.20) \\ \hline
    \end{tabular}
    \caption{Experiment of the performance of the  triggers without distillation.}
    \label{tab:abl:exp6}
\end{table}
}

\end{document}